\documentclass[10pt,letterpaper]{article}
\usepackage[top=0.85in,left=2.75in,footskip=0.75in]{geometry}

\usepackage{amsmath,amssymb}

\usepackage{changepage}

\usepackage{textcomp,marvosym}

\usepackage{cite}

\usepackage{nameref,hyperref}

\usepackage[nopatch=eqnum]{microtype}
\DisableLigatures[f]{encoding = *, family = * }

\usepackage[table]{xcolor}

\usepackage{array}

\newcolumntype{+}{!{\vrule width 2pt}}

\newlength\savedwidth

\newcommand\thickhline{\noalign{\global\savedwidth\arrayrulewidth\global\arrayrulewidth 2pt}%
\hline
\noalign{\global\arrayrulewidth\savedwidth}}

\raggedright
\usepackage[aboveskip=1pt,labelfont=bf,labelsep=period,justification=raggedright,singlelinecheck=off]{caption}

\makeatletter
\renewcommand{\@biblabel}[1]{\quad#1.}
\makeatother

\usepackage{lastpage,fancyhdr,graphicx}
\usepackage{epstopdf}
\fancyheadoffset[L]{2.25in}
\fancyfootoffset[L]{2.25in}
\microtypesetup{expansion=false}
\usepackage{underscore}

\definecolor{darkgreen}{rgb}{0,0.4,0}
\definecolor{grey}{rgb}{0.5,0.5,0.5}
\definecolor{darkgrey}{rgb}{0.3,0.3,0.3}
\definecolor{orange}{rgb}{0.7,0.4,0}

\begin{document}
\vspace*{0.2in}

\begin{flushleft}
{\Large
\textbf\newline{Approaching human parity in the quality of automated organoid image segmentation} 
}
\newline
\\
Chase Cartwright\textsuperscript{1},
Gongbo Guo\textsuperscript{3},
Sai Teja Pusuluri\textsuperscript{1},
Christopher N. Mayhew\textsuperscript{2}
Mark Hester\textsuperscript{3, 4},
Horacio E. Castillo\textsuperscript{1}
\\
\bigskip
\textbf{1} Department of Physics and Astronomy and Nanoscale and Quantum Phenomena Institute, Ohio University, Athens, Ohio 45701, USA
\\
\textbf{2} Department of Pediatrics, University of Cincinnati College of Medicine, Cincinnati Children's Hospital Medical Center, 3333 Burnet Avenue, Cincinnati, OH 45229, USA; Division of Developmental Biology, Cincinnati Children's Hospital Medical Center, 3333 Burnet Avenue, Cincinnati, OH 45229, USA; Center for Stem Cell and Organoid Medicine, Cincinnati Children's Hospital Medical Center, 3333 Burnet Avenue, Cincinnati, OH 45229, USA.
\\

\textbf{3} The Steve and Cindy Rasmussen Institute for Genomic Medicine, Abigail Wexner Research Institute at Nationwide Children's Hospital, Columbus, OH, USA.
\\
\textbf{4} Department of Pediatrics, The Ohio State University College of Medicine, Columbus, Ohio, USA 
\\

\bigskip

\end{flushleft}
\section*{Abstract}
Organoids are complex, three dimensional, self-organizing cell cultures which manifest organ-like features and represent a powerful platform for studying human disease and developing treatment options. Organoid development is characterized by dynamic morphological and cellular organization, which mimic some aspects of organ development. To study these rapid changes over the course of organoid development, advanced imaging and analytical tools are critical to accurately monitor the trajectory of organoid growth and investigate disease processes. In particular, fully automated tools allow for the rapid and consistent identification and quantification of morphological phenotypes related to genetics, growth protocol, treatment, or other variables of interest, even across large data sets. Many tools require a significant amount of task-specific training data, but recent advances in “0-shot” methods allow for image segmentation without any specific retraining.

In this work, we focus on computer vision and machine learning techniques to automatically measure the size and shape of developing spheroids derived from pluripotent stem cells (iPSCs), which are typically the starting material for generating organoid cultures. To facilitate this task, we introduce a composite method that combines the Segment Anything Model (SAM), a general-purpose foundation model, with an existing domain-specific tool. This composite method is evaluated together with several existing tools by testing them on organoid image data and comparing with the results of manual image segmentation. We find that no single existing tool is able to segment the test images with sufficient accuracy across all test conditions, but the newly introduced composite method produces consistent and accurate results for all but a very small fraction of the most challenging images. Finally, we compare the accuracy of this method to the variability between manual segmentations by independent annotators (inter-observer variability) and find that by one measure it performs at the level of inter-observer variability and by others it performs very close to it.

\section*{Author summary}
We develop several methods for automated segmentation of organoid microscopy images, and examine their performance by comparing with manually segmented images. We first consider OrganoID, a deep learning tool that segments single organoids in microscopy images, and find that in its original form it does not perform optimally with organoid image data showing misshapen organoids, presence of dead cells and/or debris, unfavorable illumination conditions, or other confounding factors. The results can be improved by either retraining OrganoID using images from related datasets, or by combining it with SAM, an extremely powerful general purpose image segmentation code. However, to reliably get accurate results we find that it is best to use a composite method that combines the retrained version of OrganoID with SAM. Those results are superior to those obtained by an extremely powerful general purpose method, namely combining Grounding DINO with SAM. A further refinement, the Hybrid method, where three alternative masks are produced (all involving SAM), and the one that has the greatest overlap with Trained OrganoID is selected, often produces the best results in absolute terms, but the improvement is usually small and may not justify the extra effort that it requires. Instead, the composite method combining Trained OrganoID + SAM is often the one that provides the best balance between costs and quality of results.

\section*{Introduction}
Organoids are three-dimensional, self-organizing assemblies of cells derived from either pluripotent or adult stem cells~\cite{Clevers2016, kretzschmar_organoids_2016}. These entities closely emulate the architecture and function of native organs, thereby providing a framework for the investigation of human biology and pathological conditions~\cite{li_organoids_2019, li_organoid_2020}. In contrast to conventional two-dimensional cell cultures, organoids comprise a variety of cell types, thereby reproducing the complexity inherent in natural tissues~\cite{hofer_engineering_2021}. Consequently, organoids are emerging tools in research domains such as organ development, pharmaceutical testing, and disease modeling. The generation of organoids entails cultivating stem cells within a controlled milieu, facilitating their differentiation and organization into intricate structures that mimic specific organs. This methodology permits the examination of cellular interactions and organ-specific functions within a regulated environment. Due to their capacity to more accurately simulate physiological conditions compared to planar cell cultures, organoids are increasingly employed in the realm of personalized medicine, permitting the assessment of pharmacological responses on a patient-specific basis \cite{Perkhofer2018, Driehuis2019, Li2020, SotoGamez2024, Abbasian2025}.

Organoids are typically analyzed through microscopy techniques. Recent advances in artificial intelligence (AI), deep learning, and image processing have yielded tools that significantly enhanced the accuracy and efficiency of this analysis~\cite{sarker_deep_2021, suganyadevi_review_2021, ker_deep_2018, moen_deep_2019, allier_frontiers_nodate}. These tools automate the detection, segmentation, and characterization of organoids in microscopy images; providing detailed morphological data such as eccentricity, solidity, area, and diameter, which are crucial for exploring the growth of organoids~\cite{broguiere_growth_2018}. These tools have been improving in their accuracy and efficiency, reducing the need for manual intervention and enabling large-scale, high-throughput analysis. Consequently, these advancements facilitate a deeper understanding of organoid biology and their potential applications in biomedical research.

There are AI tools specifically designed to work with particular types of organoids, often using tailored training data to achieve high accuracy in detecting scenarios similar to the provided data. However, these models typically struggle to generalize to other types of organoids. Notable among these tools are MOrgAna and OrgaExtractor. MOrgAna~\cite{gritti_morgana_2021} is a machine learning-based platform that quantifies and visualizes morphological data in organoid image datasets. It utilizes a supervised learning approach, usually starting with a single cropped organoid for training, to perform essential segmentation tasks for monitoring organoid development and morphological changes. Similarly, OrgaExtractor~\cite{park_development_2023} is an AI tool that enhances the accuracy and efficiency of organoid image analysis. Designed with a user-friendly interface, OrgaExtractor requires minimal image adjustments, making it accessible to researchers with limited programming expertise. Using a multi-scale U-Net architecture, it segments organoids of varying sizes from brightfield images with high accuracy and achieved an average dice similarity coefficient of 0.853~\cite{park_development_2023} on a certain test set. It is particularly advantageous for analyzing organoids with fluorescence reporter systems during their developmental stages, and it correlates well with the proliferation assay, accurately reflecting actual organoid culture conditions and aiding in determining optimal subculture times~\cite{park_development_2023}. 

OrganoID~\cite{matthews_organoid_2022} is another deep learning tool designed for segmenting organoids in brightfield and phase-contrast microscopy images. It provides automatic recognition, labeling and tracking capabilities for organoids within a single image. Its segmentation process operates at the pixel level, enhancing the precision of detection and characterization. It generates both complete masks and a suite of metrics such as organoid area, eccentricity, and solidity. It is also fully trainable using task-specific, manually generated segmentation data. We have utilized OrganoID as a baseline in several experiments, leveraging its segmentation features to compare and benchmark against other methodologies. Further details on the implementation and results of using OrganoID in our study are provided in later sections.

AI models such as U-Net, the Segment Anything Model (SAM), and Grounding DINO, are trained on millions of images, allowing them to generalize effectively across various scenarios. These are often refered to as Zero-Shot methods, as they require no training data to perform segmentation. U-Net, a well-known convolutional neural network (CNN) architecture, is extensively used for biomedical image segmentation~\cite{ronneberger_u-net_2015}. It features an encoder-decoder structure that captures both spatial context and high-resolution details. The U-Net architecture has evolved into several variants, such as the multi-scale U-Net, which improves performance through multi-scale feature learning and residual paths~\cite{su_frontiers_nodate}. This architecture enables precise organoid segmentation, even in complex and noisy microscopy images. In parallel, SAM is a foundation model trained on a vast dataset of 11 million images across diverse modalities~\cite{kirillov_segment_2023}. It automates the detection of individual objects by employing comprehensive post-processing steps to enhance segmentation results, ensuring accurate identification. SAM effectively resolves common challenges like background misidentification and can be adapted to microscopic images~\cite{archit_segment_2025}, establishing itself as a robust tool for detecting organoids under varied imaging conditions. Grounding DINO is an open-set object detector that generates bounding boxes from an image and a short free-form text prompt, with remarkable accuracy and often without the need for domain-specific training~\cite{liu_grounding_2024}. It is often referred to as a “describe-and-detect” engine. It does not perform full image segmentation, but it can serve as a flexible companion to pixel-level segmentation engines such as SAM. In order to obtain image segmentation starting directly from a text prompt, approaches that use Grounding DINO combined with SAM have been tried~\cite{RenGroundedSAM2024, Mumuni2024}, but never, to our knowledge, in the context of organoid analysis.

The integration of AI tools has revolutionized the field of organoid research, providing accurate and automated solutions for the detection and analysis of organoids. These advancements facilitate high-throughput studies and significantly advance our understanding of organ development, disease mechanisms, and drug responses. However, despite the progress brought about by AI-based tools for organoid detection, several challenges and drawbacks persist. Understanding these limitations is crucial for further refinement and development of more robust and versatile tools.

One of the primary drawbacks of current methods is their reliance on manual image analysis and annotation, which is labor-intensive, time-consuming, and prone to human error. Even with advanced tools like OrganoID, MOrgAna and OrgaExtractor, the initial stages often require extensive manual intervention to create training datasets~\cite{matthews_organoid_2022, gritti_morgana_2021, park_development_2023}. This not only slows down the research process but also introduces variability due to subjective interpretations of organoid morphology. Most supervised deep learning-based tools, such as OrganoID and OrgaExtractor, require substantial annotated data for training and may need retraining for different types of microscopy images, limiting their generalizability. Each distinct microscopy modality may necessitate a new set of annotations and model adjustments, making it challenging to apply these tools across various experimental conditions and organoid types without significant customization.

Changes in organoid morphology can indicate important phenotypic responses and state transitions, such as the epithelial-mesenchymal transition in tumor models. Existing methods often struggle to capture these complex morphological changes accurately, particularly in heterogeneous organoid populations~\cite{schuster_automated_2020}. For example, some tumor organoids grow into structures with invasive projections into the culture matrix, reflecting significant biological processes. Current tools may not always detect these nuanced changes reliably. While tools like OrganoID offer automated profiling of individual organoid morphology, they face challenges in accurately quantifying complex shape metrics. Metrics such as circularity, solidity, and eccentricity are critical for understanding phenotypic responses to treatments like chemotherapeutic agents. These metrics can be affected by the quality of segmentation and the precision of contour detection, potentially leading to misleading conclusions about the effects of treatments on organoid morphology~\cite{park_development_2023}.

Another significant drawback is the integration and interpretation of data from different sources and modalities. Combining data from various imaging techniques (e.g., brightfield, fluorescence) and ensuring consistent analysis across these datasets can be challenging. Tools like SegmentAnything and OrgaExtractor provide solutions for specific imaging types, but integrating results from multiple sources into a coherent analysis framework remains complex~\cite{matthews_organoid_2022}. There is also a lack of standardized protocols and benchmarks for evaluating the performance of different organoid analysis tools. This makes it difficult to compare results across studies and hampers the reproducibility of research findings~\cite{park_development_2023}. Establishing standardized datasets and evaluation metrics would be beneficial for advancing the field and ensuring the reliability of AI-based organoid analysis methods.

While Zero Shot models like SegmentAnything offer a degree of generalizability without the need for extensive retraining, addressing the existing limitations of current methods is essential for their broader adoption and more effective application in biomedical research. Continuous improvement in algorithm robustness, generalizability, and integration capabilities will pave the way for more precise and efficient organoid analysis.

In this study, we combined several of these tools, namely OrganoID, GroundingDINO, and SegmentAnything, to segment microscopy images of iPSC-derived spheroids under less than ideal conditions and extracted size and shape parameters from those segmentations. In total, we assembled 7 different image segmentation procedures using these 3 tools, and we compared their accuracy. To do this, we trained OrganoID using a selection of training images of spheroids from 11 iPSC cell lines, and then selected 3 sets of images with diverse morphological features as test data to benchmark our segmentation methods. After manually segmenting these 3 sets of images, we compared each of the 7 segmentation procedures to these manual results and found that a composite method which combines OrganoID and SegmentAnything can provide better results than any individual method and also outperforms other combinations of tools.

\section*{Materials and methods}
\label{sec:MaterialsMethods}
\subsection*{Cell cultures and images}
\label{sec:CellCulturesImages}
\begin{figure}[h]
\includegraphics[width=\textwidth]{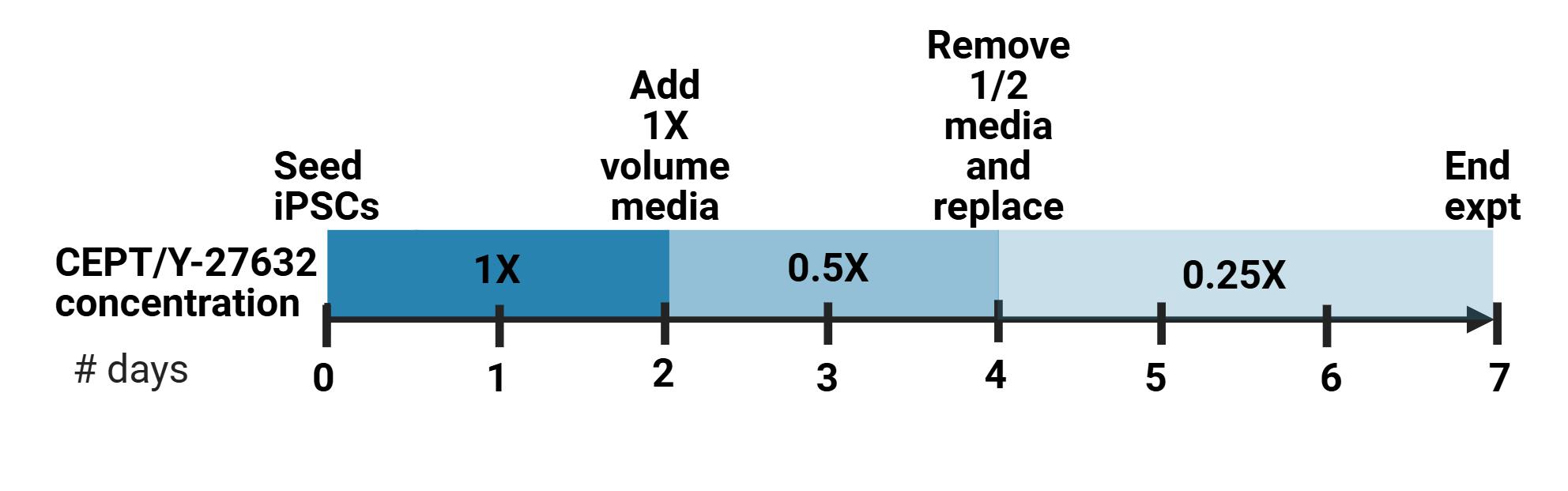}
\caption{Growth protocol for iPSC-derived spheroids. Media was replaced in well plates as shown.}
\label{fig:growthprotocol}
\end{figure}

Eleven independent healthy, control iPSC lines, or iPSC lines either genetically modified with reporter genes or CRISPR gene corrected iPSCs were cultured in StemFlex medium (ThermoFisher, A3349401) (Table 1) for these experiments. In brief, iPSC lines were passaged at least twice and then dissociated with TryPLE (GibcoTM, 12604013) for approximately five minutes and then neutralized with StemFlex medium. Single iPSCs were seeded in non-tissue culture, untreated u-bottom 96-well plate and seeded at 8,000 cells per well in StemFlex medium alone (untreated); or containing either 5µM Y-27632 ROCK inhibitor (Selleckchem, S1049), or 1X CEPT (FUJIFILM, NC2668999), (both according to the manufacturer’s instructions); or containing a reduced CEPT concentration (1/2, 1/4, 1/10 or 1/20 of the recommended value). Plates were then balanced and cells were centrifuged at 400g for 4 minutes to collect cells within the middle of each well. Phase contrast microscopic images were captured from all 96 wells per dish and conditions every day for seven days using an EVOS M7000 Imaging System (ThermoFisher). Media was replaced in 96 well plates according to the schematic in Fig.~\ref{fig:growthprotocol}. 

\begin{table}[!ht]
\begin{adjustwidth}{-2.25in}{0in} 
\centering
\begin{tabular}{| c + c | c | c | m {6cm} |}
	\hline
	\textbf{Number} & \textbf{Source} & \textbf{Cat. Number/ID} & \textbf{Sex}\\
	\thickhline
	\textbf{iPSC-1} & ThermoFisher & A18945 & female \\
	\hline
	\textbf{iPSC-2} & ALSTEM,LLC & iPS15 & N/D \\
	\hline
	\textbf{iPSC-3} & Allele Biotechnology&	ABP-SC-HDFAIPS&	male \\
	\hline
	\textbf{iPSC-4} & Allen Cell Collection/Corielle Repository 	&iAICS-0036-028 CAG-eGFP 	&male	\\
	\hline
	\textbf{iPSC-5$^*$} & Allen Cell Collection/Corielle Repository &	AICS-0078 cl.79 TOMM20-eGFP / TUBA1B-RFP &	female	\\
	\hline
	\textbf{iPSC-6} & Applied StemCell  &	TARGATT™ hiPSCs &	female\\
	\hline
	\textbf{iPSC-7} & Allen Cell Collection/Corielle Repository &	AICS-0023 cl.20 TJP-eGFP &	male\\
	\hline
	\textbf{iPSC-8} & Cincinnati Children's Hospital/Mayhew Lab & iPSC_75.1 & male\\
	\hline
	\textbf{iPSC-9$^*$} & Cincinnati Children's Hospital/Mayhew Lab&	iPSC_115.1&	male \\
	\hline
	\textbf{iPSC-10$^*$} & Cincinnati Children's Hospital/Mayhew Lab&	iPSC_72.3&	male\\
	\hline
	\textbf{iPSC-11}$^{**}$ & Mass. General Hospital/Ramesh Lab  & TSC1 Isogenic Control & N/D \\
	\hline

\end{tabular}
\end{adjustwidth}
\caption{
{\bf Cell lines used in training and test data} 
iPSC lines used to generate training and test data for image segmentation. Images were taken from spheroids derived from these cell lines under a variety of culture conditions (see Section 'Materials and Methods' for details). All iPSC lines were used to generate training data. {}$^*$: Cell lines used to generate test data (see Subsection 'Test Data'). {}$^{**}$: Cell line provided by Dr. Vijaya Ramesh from Harvard Medical School, Center for Genomic Medicine, Massachusetts General Hospital.
\label{cell_lines}}
\end{table}

\subsection*{Training data}
\label{sec:TrainingData}
To generate a robust and diverse dataset to retrain OrganoID, we used 176 manually segmented and labeled images taken from a variety of experimental conditions (either untreated, treated with CEPT at standard concentration, treated with CEPT at 1/2, 1/4, 1/10 or 1/20 of standard concentration, or treated with Y-27632) across the 11 cell lines listed in Table 1, and across different time points. 

We chose the first 70 images at random, segmented them manually, and used them to retrain the OrganoID model weights, calling this Trained OrganoID version 1. For version 2, we selected a representative sample of images that version 1 segmented poorly, segmented them manually, added them to the training data, and retrained the model weights again. We repeated this until version 6, where the performance was only marginally better than version 5, and decided that adding more training data was not the most effective way of further increasing accuracy. Unless otherwise specified, any further reference to Trained OrganoID will refer to the version 6 model weights.

\subsection*{Test data}
\label{sec:TestData}
\begin{figure*}[h!]

\begin{adjustwidth}{-2.25in}{0in} 
\centering

\includegraphics[width=1.4\textwidth]{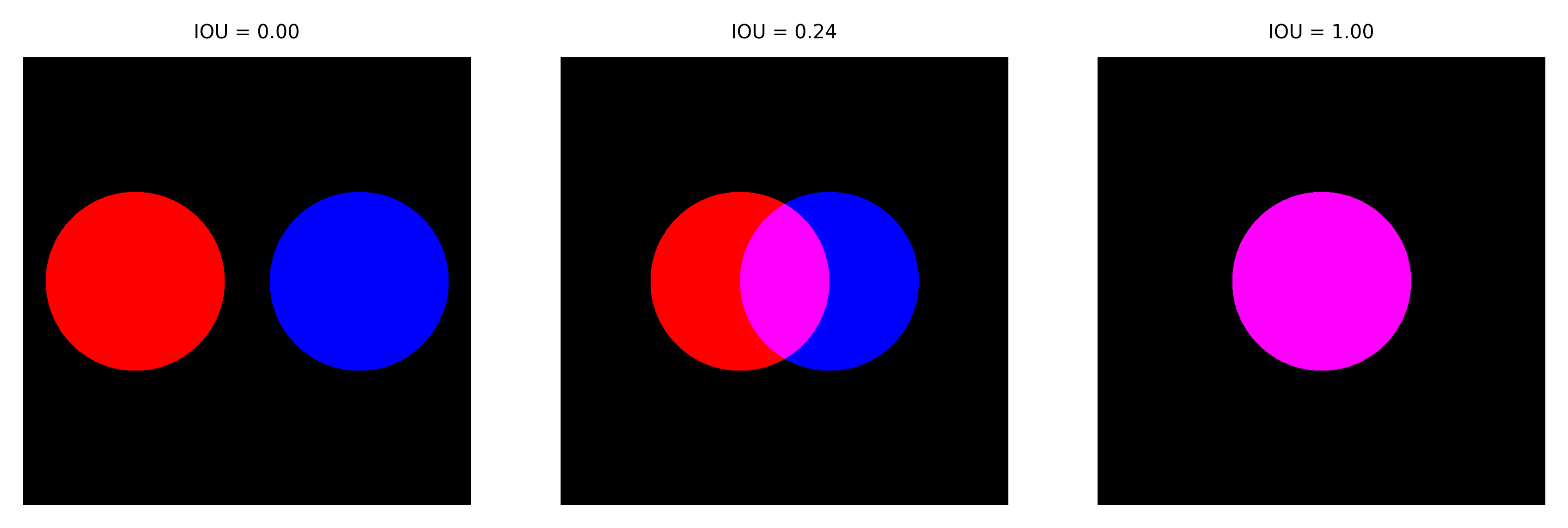}
\caption{{\bf Demonstration of IOU Calculation} We calculate IOU by dividing the number of pixels which belong to both masks (the purple region) by the total number of pixels which appear in either mask (the blue, purple, and red regions combined) in order to quantify the agreement of two masks.}
\label{fig:ioudef}
\end{adjustwidth}
\end{figure*}

In order to assess our segmentation methods effectively, we utilized test data that would be challenging to segment, based on diverse morphologies observed in iPSC spheroids. Previous reports have demonstrated that a cocktial of the following small molecules: Chroman 1, Emricasan, Polyamine Supplement, and Trans-ISRIB, referred to as CEPT, signficantly enhance the survival of iPSCs under stressful conditions~\cite{Chen2021}. At the recommended concentration of CEPT, all iPSC lines formed spheroids, showing high viability and minimal cell loss, and every method we evaluated could consistently and accurately segment those spheroids. In order to generate test data in which we could rigorously evaluate our analytical tools, we generated morphologically diverse spheroids across iPSC lines by reducing the CEPT concentration to 1/10 or 1/20 of the recommended value. The images thus produced were grouped into three sets, which we label as {\bf Image Set A}, {\bf Image Set B} and {\bf Image Set C}. GG and CC then worked together to create manually segmented masks which we can use as the ground truth for comparison. To quantify the similarity between two masks, we used the intersection-over-union (IOU), which is defined as the ratio between the number of pixels common to both masks and the number of pixels included in either mask, as shown in Fig.~\ref{fig:ioudef}.

\begin{figure}[h!]
\begin{adjustwidth}{-2.25in}{0in} 
\centering
\includegraphics[width=1.45\textwidth]{"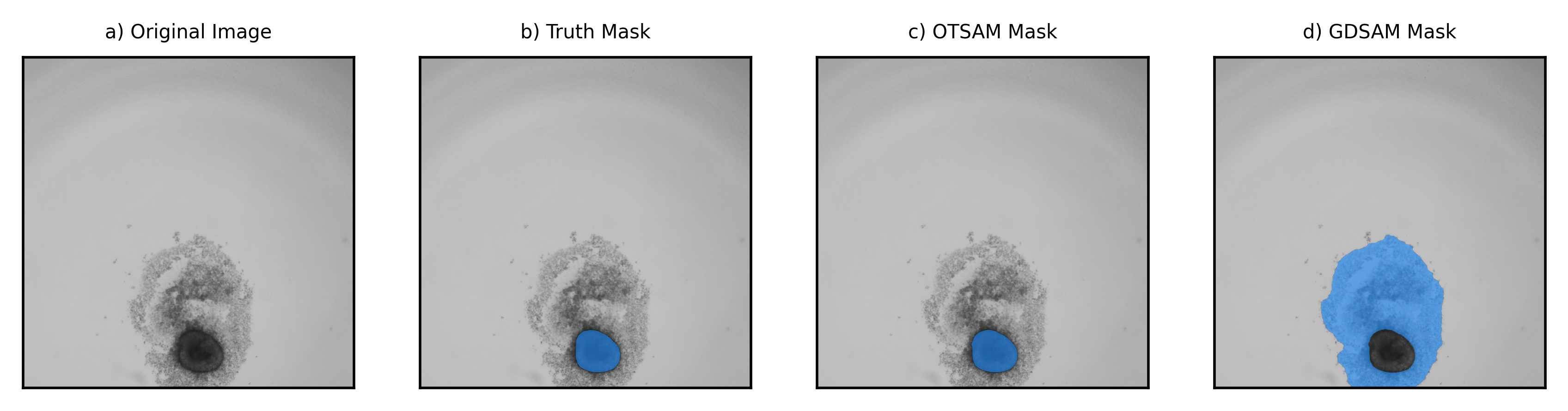"}
\caption{{\bf Example image and masks from Image Set A} One of the 48 images in {\bf Image Set A} with sample masks overlaid for demonstration. a) The original image which the computational tools take as input. b) A manually generated mask that we treat as the ground truth. c) The mask produced by the Trained OrganoID + SAM Composite method, which is almost indistinguishable from the ground truth mask by visual inspection. (IOU = 0.98) d) The mask produced by the Grounding DINO + SAM method, which is almost disjoint with the ground truth mask. (IOU = 0.00)}
\label{examplea}
\end{adjustwidth}
\end{figure}

{\bf Image Set A} consists of 48 images taken on Day 4 of spheroids grown from the iPSC-5 cell line, treated with a 5\% fraction of the recommended CEPT concentration. In this image set, there is a large amount of dead cells and debris surrounding the spheroids, which makes the edges difficult to distinguish. Fig.~\ref{examplea}a shows an example image from this set, Fig.~\ref{examplea}b shows the manually generated mask, which is the benchmark to which we will compare the automated methods, Fig.~\ref{examplea}c shows an example of a good automatically generated mask, which is indistinguishable to the naked eye from the manual one, and Fig.~\ref{examplea}d shows an extremely poor automatic segmentation which is almost completely disjoint with the true spheroid.

\begin{figure}[h!]
\begin{adjustwidth}{-2.25in}{0in} 
\centering
\includegraphics[width=1.45\textwidth]{"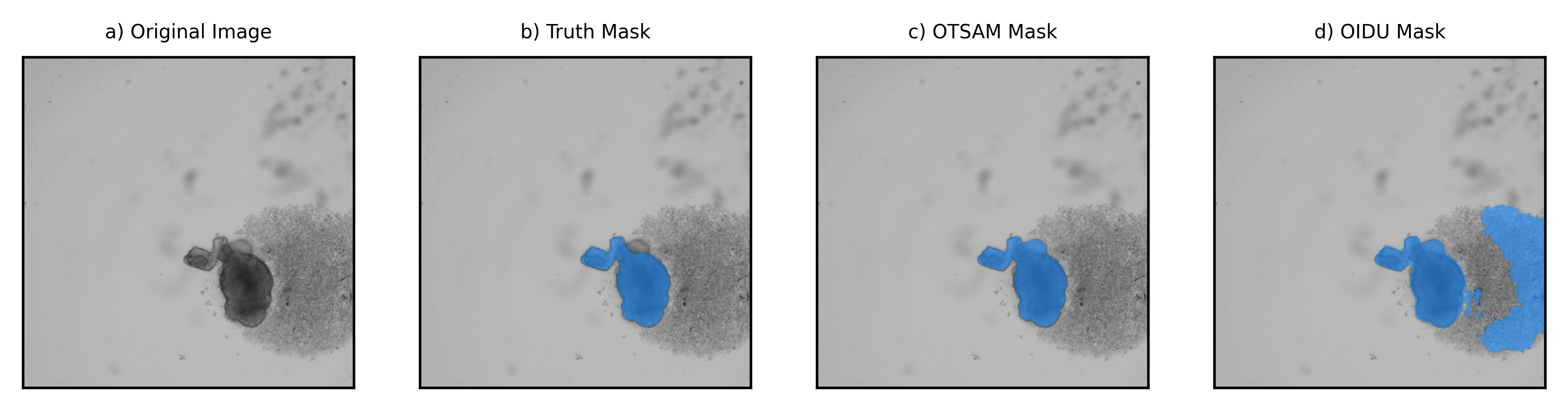"}
\caption{{\bf Example image and masks from Image Set B} One of the 48 images in {\bf Image Set B} with sample masks overlaid for demonstration. a) The original image which the computational tools take as input. b) The manually-generated ground truth mask. c) The Trained OrganoID + SAM Composite mask, indistinguishable from the ground truth except for a small protrusion on the upper-right edge of the organoid. (IOU = 0.93) d) The Untrained OrganoID mask which contains the actual organoid as well as a large section of the debris cloud on the left. (IOU = 0.35)}
\label{exampleb}
\end{adjustwidth}
\end{figure}

{\bf Image Set B} contains 48 images taken on Day 5 of spheroids from the iPSC-9 cell line, treated with 10\% of the recommended CEPT concentration. In this image set, there is still a large quantity of debris around the spheroids, but the spheroids themselves have extremely irregular shapes, so many segmentation methods will erroneously include pieces of debris or exclude protrusions of the spheroid itself. As in Fig.~\ref{examplea}, Fig.~\ref{exampleb} shows (a) the original, raw image, (b) the correct, desired mask, (c) an automatically generated mask which is very close to the correct one, and (d) an example of a poor segmentation. 

\begin{figure}[h!]
\begin{adjustwidth}{-2.25in}{0in} 
\centering
\includegraphics[width=1.45\textwidth]{"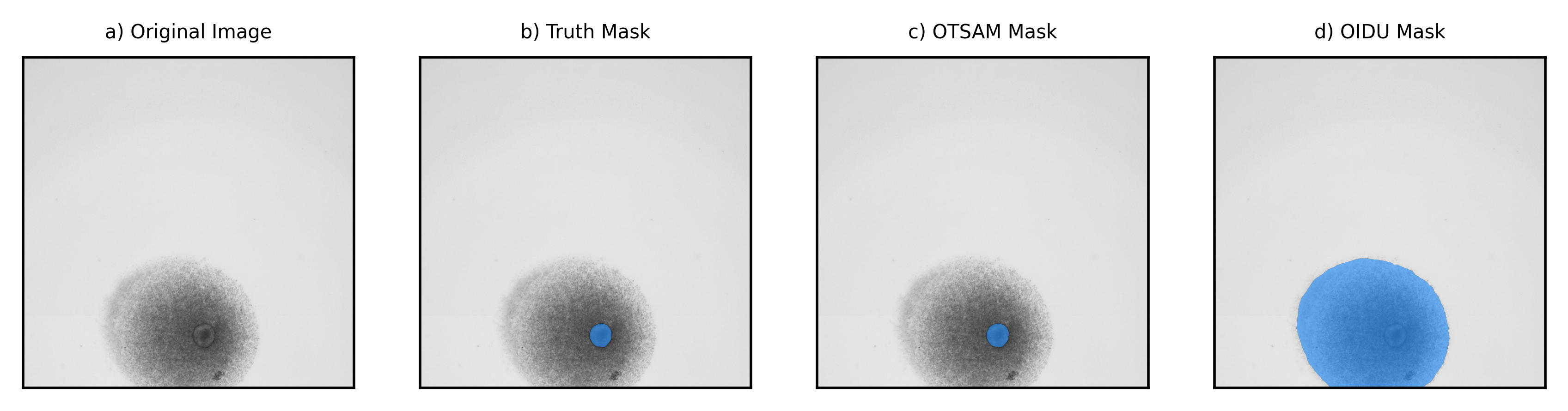"}
\caption{{\bf Example image and masks from Image Set C} One of the 48 images in {\bf Image Set C} with sample masks overlaid for demonstration. a) The original image which the computational tools take as input. b) The manually-generated ground truth mask. c) The mask produced by the Trained OrganoID + SAM Composite method, which is almost indistinguishable from the ground truth mask by visual inspection. (IOU = 0.97) d) Grounding DINO + SAM produces a mask that covers both the organoid and the much larger surrounding debris cloud. (IOU = 0.03)}
\label{examplec}
\end{adjustwidth}
\end{figure}

{\bf Image Set C} contains 48 images taken on Day 2 from the iPSC-10 cell line, treated with 5\% of the recommended CEPT concentration. This is, by far, the most difficult image set to segment among the three. These images are taken so early in the process of spheroid growth that the iPSCs have only begun to integrate, so the spheroid is not clearly distinguished from the surrounding cloud of unintegrated cells. The spheroid is sometimes lighter than the surrounding cloud, which makes it even more difficult to segment automatically. Fig.~\ref{examplec} shows (a) an example image, (b) the desired mask, (c) a very good automatically generated mask, and (d) a typical erroneous mask for this image set, where the whole cloud of cells is mistaken for a spheroid. 

On average, {\bf Image Set A} contains the easiest images to segment, and {\bf Image Set C} contains the hardest, and the images within sets are ordered similarly, in descending order of the average intersection-over-union (IOU) across all automated segmentation methods, as seen in Fig.~\ref{fig:all}a.

\subsection*{Computational methods}
The computational tools used in this work are OrganoID~\cite{matthews_organoid_2022},  Grounding DINO~\cite{liu_grounding_2024}, and the Segment Anything Model (SAM)~\cite{kirillov_segment_2023}, all three of them already discussed in the Introduction. 
We used two versions of OrganoID: one with the original, pretrained model weights, and another with retrained weights derived from our own data. For the retraining we used 176 manually-segmented images from our own experiments (disjoint from the Image Sets used for testing, see~``Training Data''~\ref{sec:TrainingData} above).
Grounding DINO gives a way to detect the location of spheroids without any training data, provided that we can find a sufficiently clear text prompt. However, since it is trained and benchmarked on general-purpose image libraries and annotation data, not on microscopy images, we cannot use technical language in the prompt. With that in mind, we have used the prompt, “A dark, solid cluster.” Among the prompts we considered, this one most consistently identified spheroids at every stage of development. 
For running SAM, we used all three allowed input modalities: either giving a point (generated by OrganoID) within the desired object, or giving a bounding box (generated by Grounding DINO) that contains the desired object, or providing no information, in which case SAM generates a collection of masks corresponding to every detected object in the image.

By combining the above tools, we proposed the following seven methods to generate complete masks for the test images: i) Untrained OrganoID (OIDU), running OrganoID with its original, pretrained model weights; ii) Grounding DINO + SAM (GDSAM), creating a bounding box with Grounding DINO and using it as input for SAM; iii) Untrained OrganoID + SAM (OUSAM), allowing SAM to create masks for all detected objects in the image, and then selecting and stitching together those mask whose overlap with the OIDU mask is higher than a fixed threshold; iv) Trained OrganoID (OIDT), using OrganoID with retrained weights derived from our own data; v) Trained OrganoID + SAM (OTSAM), similar to Untrained OrganoID + SAM but using the trained version of OrganoID; vi) OrganoID Centroid + SAM (OCSAM), using the centroid of each organoid computed by the trained version of OrganoID as the input for SAM; and vii) Hybrid method, a combination of the above (discussed later). The first three methods require no training, and are thus much simpler to implement. The other four methods require training OrganoID, and are therefore more demanding in their implementation, but have a higher potential for giving accurate results. The results from all seven methods are compared with a ground truth consisting of manually generated masks. 
  
For all of the methods for mask generation, including the ground truth, the metrics, such as organoid area, eccentricity, and solidity, were calculated with code adapted from OrganoID using the "skimage" Python library~\cite{VanDerWalt2014}.

\section*{Results and discussion}

\begin{figure}[!h]
\begin{adjustwidth}{-2.25in}{0in} 
\centering
\includegraphics[width=1.4\textwidth]{"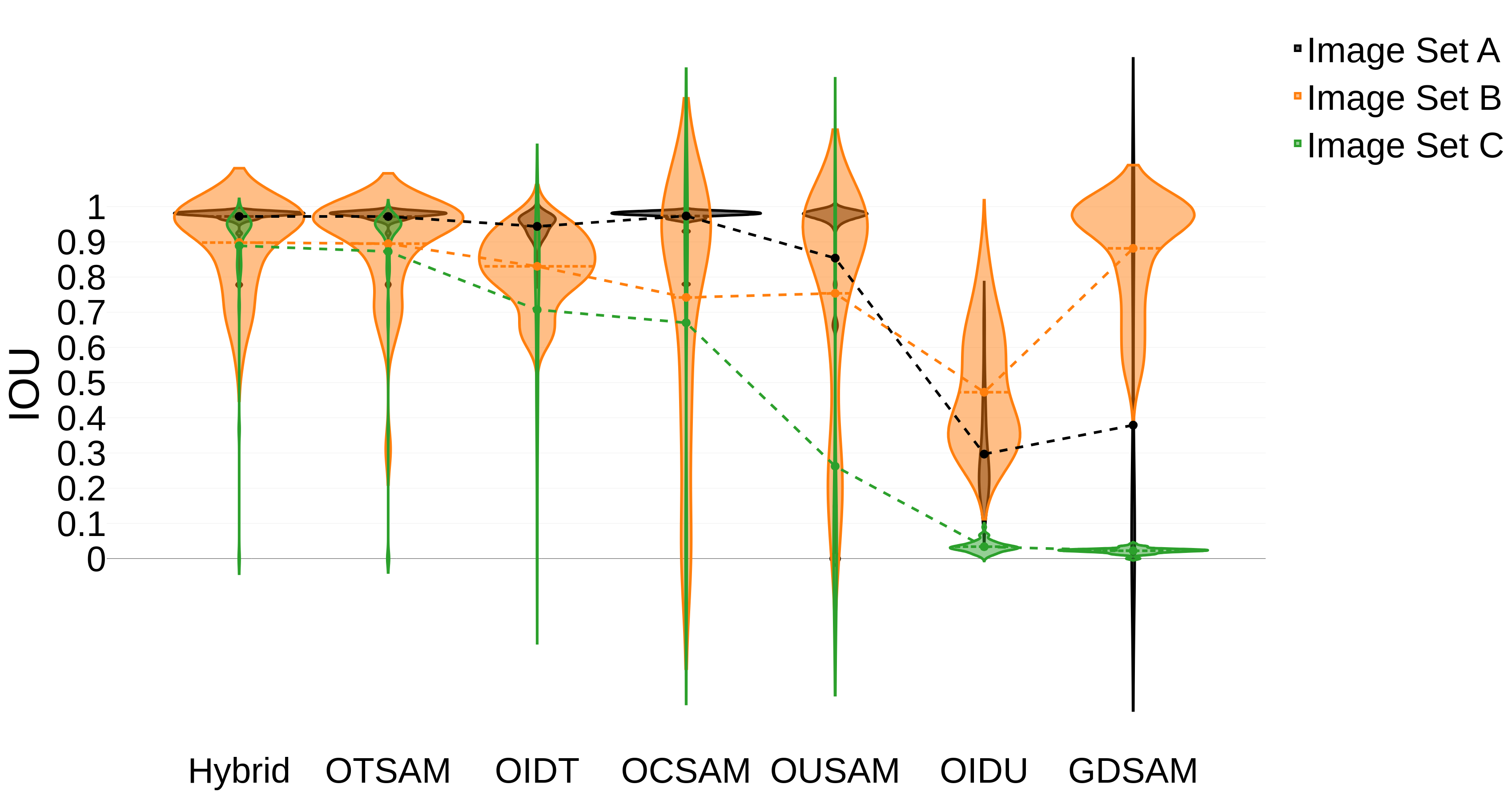"}
\caption{{\bf IOUs of Segmentation Methods}
IOUs between the masks produced by each method and the corresponding ground truth masks. On the horizontal axis is the segmentation method used to produce the masks and on the vertical axis is the overlap between those masks and the corresponding ground truth masks, which ranges between 0 and 1, with 1 being perfect accuracy. Dotted lines indicate the mean IOU for the respective image set and segmentation method.}
\label{meanious}
\end{adjustwidth}
\end{figure}

In this section we report the results of applying the different segmentation methods we proposed. An overall picture of the results is given by Fig.~\ref{meanious}, which  shows the mean IOUs with the manually segmented masks for each segmentation method and image set.

\begin{figure}[!h]
\centering
\includegraphics[width=.9\textwidth]{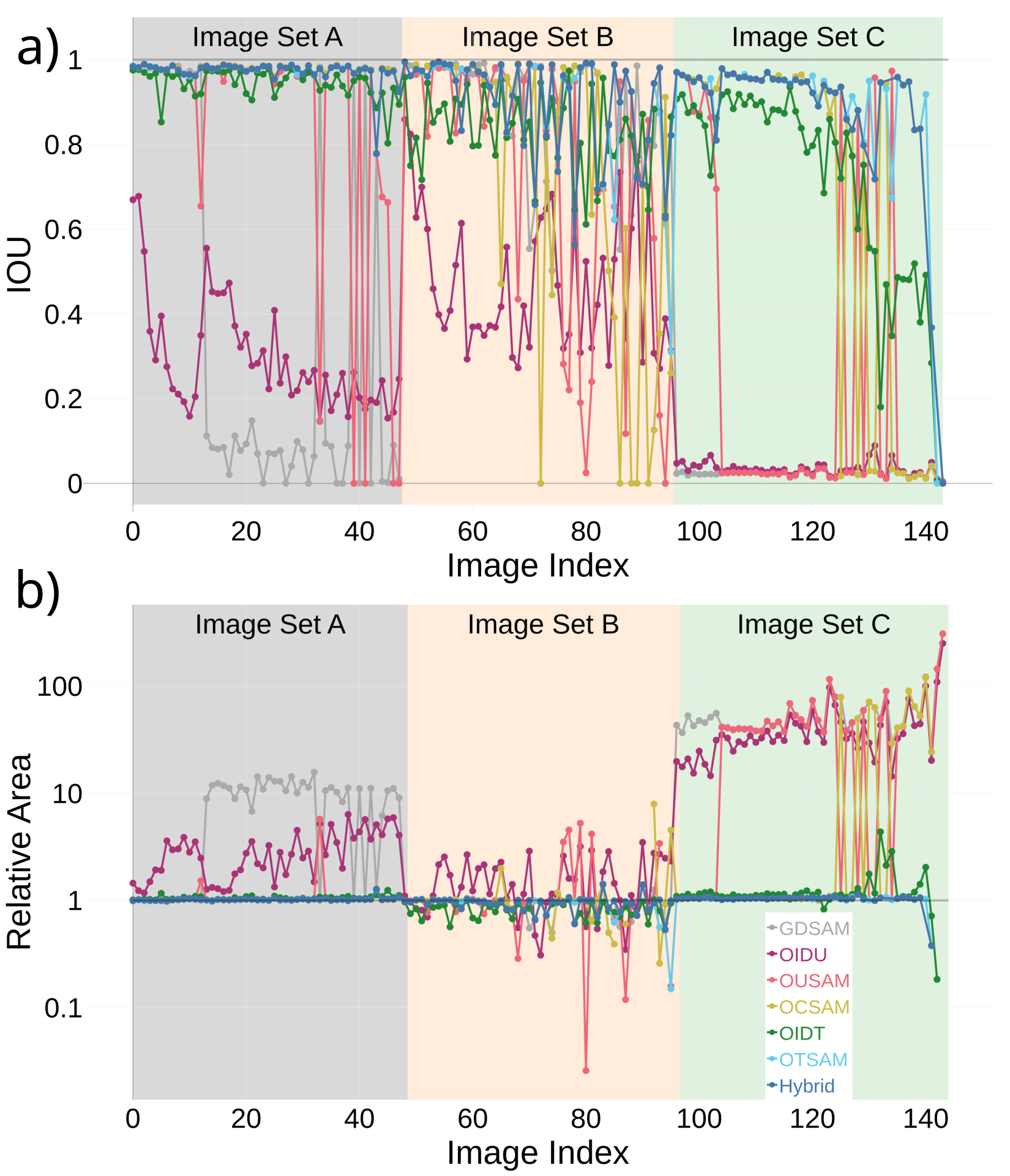}
\caption{{\bf IOUs and normalized areas for each image and segmentation method} a) IOU (as defined in Fig.~\ref{fig:ioudef}) between each mask and the ground truth mask for the same image. The results for each of the 144 images across the 3 image sets are shown. Note that different segmentation methods perform differently on different image sets. The images are sorted within image sets so that, on average, the IOU is decreasing as the image index increases. The image index is shown on the horizontal axis with the IOU on the vertical axis and the line color represents the segmentation method used to produce each mask. The legend is shown in panel b. b) Relative area, i.e.~ratio of the area of one particular organoid as measured by a particular segmentation method to the ground truth area of the same organoid. The horizontal axis and line colors represent the same things as in panel a. The legend shown here also applies to Figs.~\ref{GD}, \ref{fig:OID}, \ref{traineddata}, \ref{untrained}, \ref{hybrid}, \ref{big4}, and \ref{big4cont}, each of which shows results for a subset of the 7 segmentation methods. For all segmentation methods except the Hybrid method a missing data point represents an area of 0 which cannot be plotted on a logarithmic scale. The Hybrid method alone has a handful of missing data points in {\bf Image Set C} where it was not sufficiently confident to return a mask.}
\label{fig:all}

\end{figure}

A more detailed picture is provided by Fig.~\ref{fig:all}, where results are shown for the segmentation of individual images. Fig.~\ref{fig:all}a shows the IOU of each mask with the ground truth mask of the same image. (Of course, the IOU of the ground truth mask with itself is 1 by definition, and the ratio of the area of the ground truth mask to itself is also 1. To avoid cluttering figures, trivial ground truth data such as those are excluded from most plots.)
Fig.~\ref{fig:all}b shows the ratio of each measured area to the actual, ground truth area for each image. This shows whether the particular segmentation method has a significant bias, either to create a mask far larger than the actual organoid or to only recognize a small piece of it. We define "relative area" or "normalized area" as the ratio of the area measured for the mask in question to the area of the ground truth mask for that same spheroid. (If multiple spheroids are identified, only the largest is considered.).

\begin{figure}[h!]
\begin{adjustwidth}{-2.25in}{0in} 
\centering

\includegraphics[width=1.4\textwidth]{"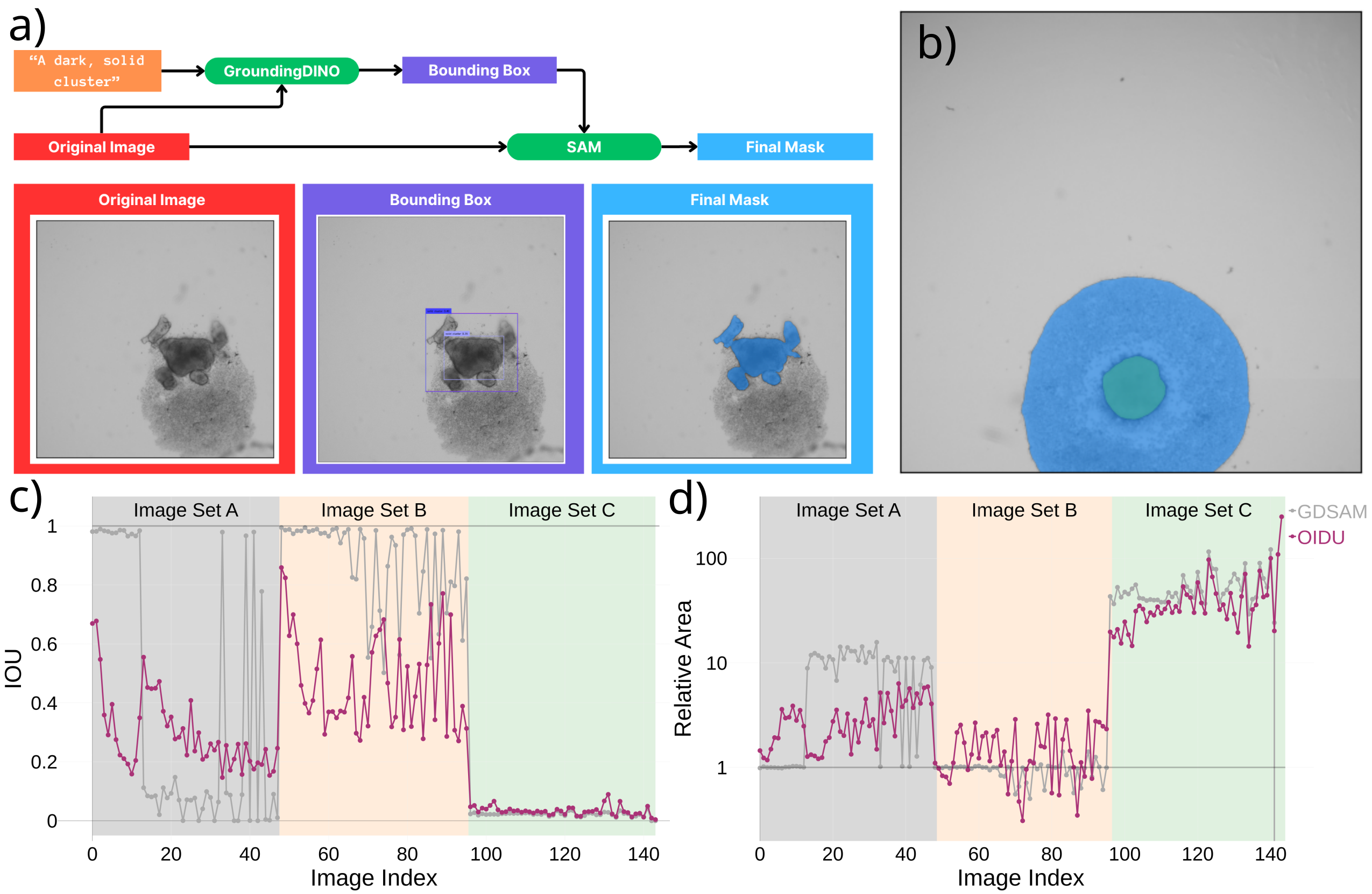"}

\caption{{\bf Untrained Segmentation Processes} a) Flowchart for the Grounding DINO + SAM process: Grounding DINO takes the original image along with the text prompt {\it "a dark, solid cluster"}, and outputs bounding boxes representing regions that may contain an object fitting the given description. The Segment Anything Model then takes the original image and the Grounding DINO bounding boxes, generating masks for the objects that best correspond to the given boxes. b) An example mask (blue) produced by Grounding DINO + SAM from an image in {\bf Image Set A} compared to the ground truth mask (green). The masks are very different (IOU=0.08). c) IOUs for the Grounding DINO + SAM (GDSAM) and Untrained OrganoID (OIDU) methods. selected from Fig.~\ref{fig:all}a. Grounding DINO + SAM does better on {\bf Image Set B}, but worse on the other two image sets. Grounding DINO + SAM also tends to generate masks with very high or very low IOUs, a trend we will also observe with other methods using SAM. d) Relative areas for Grounding DINO + SAM and Untrained OrganoID methods, selected from Fig.~\ref{fig:all}b. Both methods tend to overestimage the true areas on {\bf Image Sets A and C}, but Grounding DINO + SAM tends to err more dramatically, especially on {\bf Image Set C}.}
\label{GD}
\end{adjustwidth}
\end{figure}

\begin{figure}[!h]
\begin{adjustwidth}{-2.25in}{0in} 
\centering
\includegraphics[width=1.4\textwidth]{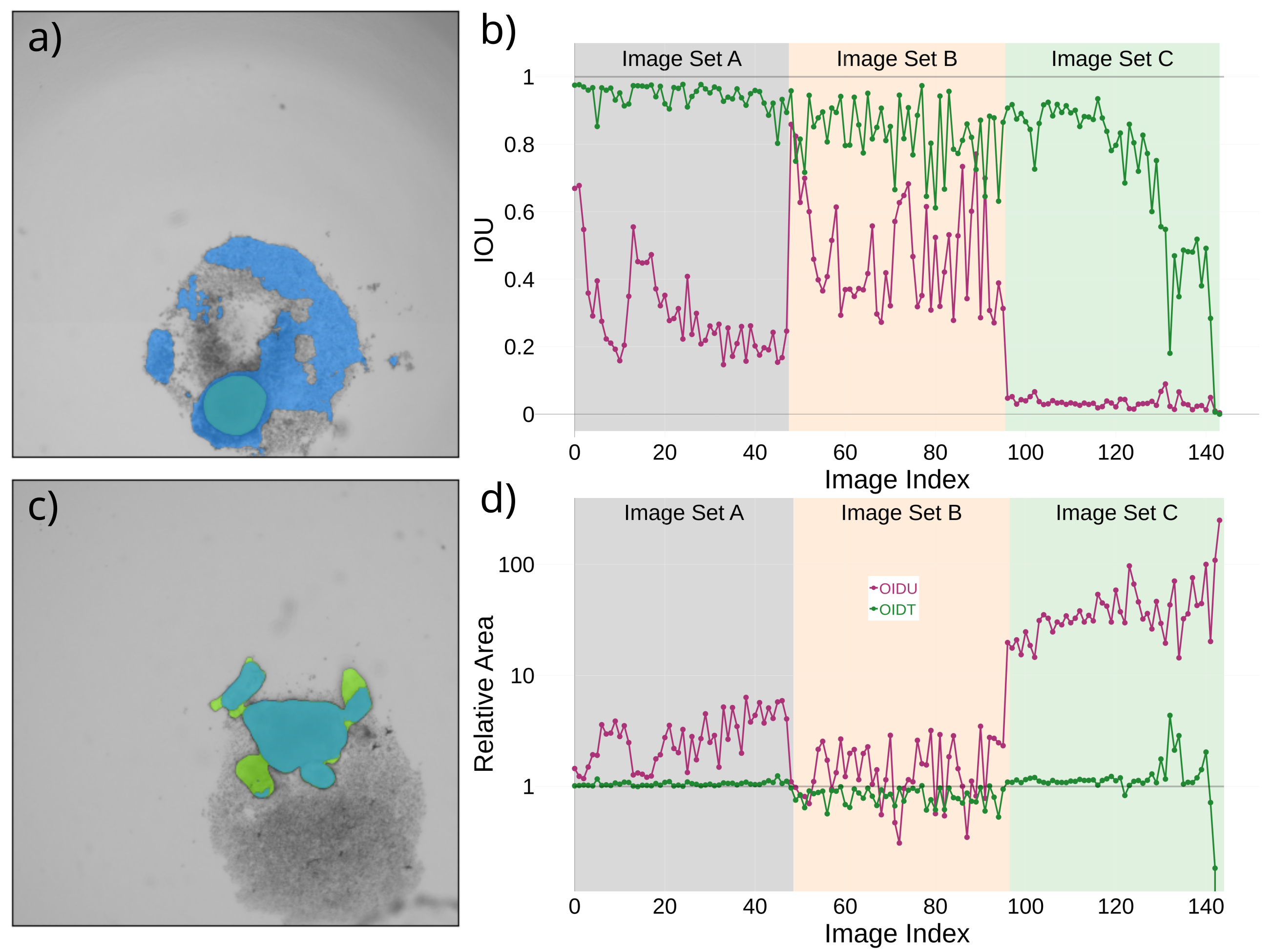}

\caption{{\bf IOUs and normalized areas for Trained and Untrained OrganoID} a) An example mask produced by Untrained OrganoID (blue) from an image in {\bf Image Set A}, compared to the ground truth mask (green). (IOU=0.15). b) IOUs for the Trained (OIDT) and Untrained OrganoID (OIDU) methods, selected from Fig.~\ref{fig:all}a. Trained OrganoID performs dramatically better, not only on average for each image set, but also for almost every individual image. The improvements are especially noticeable on {\bf Image Sets A and C}. c) An example mask produced by Trained OrganoID from an image in {\bf Image Set B}. (IOU=0.80) d) Relative area for the Trained and Untrained OrganoID methods, selected from Fig.~\ref{fig:all}b. Not only is Trained OrganoID far more accurate on every image set, but it also shows very little bias toward overestimating or underestimating area, unlike Untrained OrganoID, which consistently overestimated the areas of organoids, especially on {\bf Image Set C}.}
\label{fig:OID}
\end{adjustwidth}
\end{figure}

In order to segment our images, we first consider two of the methods that require no training: Untrained OrganoID (OIDU) and Grounding DINO + SAM (GDSAM). OrganoID is pretrained to identify and segment organoids, while Grounding DINO + SAM is a common combination of tools used for general segmentation tasks. Grounding DINO takes a text prompt and generates bounding boxes around potential objects matching the prompt, while SAM takes the bounding boxes and segments the exact boundary of the object contained therein (Fig.~\ref{GD}a). As seen in Fig.~\ref{meanious} and Fig.~\ref{GD}c, OrganoID is more consistent and does slightly better on 2 out of 3 image sets, but never performs particularly well, whereas Grounding DINO + SAM is much more likely to be either exactly correct or very incorrect. Fig.~\ref{fig:OID}a shows that OrganoID usually finds the organoid itself, but is likely to either cut off sections or to extend beyond the actual organoid in unpredictable ways. Grounding DINO + SAM, on the other hand, will often segment the entire cloud of dead cells and debris around the actual organoid, which causes it to consistently overshoot area measurements on some image sets, as seen in Figs.~\ref{GD}b and~\ref{GD}d. As indicated before, the text prompt we found most effective for Grounding DINO + SAM is “a dark, solid cluster”, so it’s not surprising that this method does particularly poorly on {\bf Image Set C}, where the actual organoids are sometimes lighter than the surrounding debris cloud (Fig.~\ref{examplec}). While Grounding DINO + SAM does fairly well on some images and might work well for some applications, neither it nor OrganoID work well enough on our data. While there is no obvious way to improve Grounding DINO + SAM, except perhaps discovering a better prompt, there are several ways we can enhance OrganoID.
	
The first and most obvious way to make OrganoID more accurate is to retrain it for our own data sets. OrganoID has a built-in retraining protocol, and we applied it by using 176 manually segmented images, as discussed before. Trained OrganoID (OIDT) dramatically outperforms the untrained version on every image set, as seen in Figs.~\ref{fig:OID}b and~\ref{fig:OID}d, but it still tends to make small mistakes around the boundaries of the organoid (Fig.~\ref{fig:OID}c). We saw with Grounding DINO + SAM that SAM is very good at finding the boundaries of objects and following them exactly, but Grounding DINO was struggling to identify the organoids, so the method would generate a mask that was either exactly right or completely wrong. Trained OrganoID, on the other hand, is very consistently identifying the organoid, but failing to follow the boundary precisely, so combining SAM with OrganoID instead of Grounding DINO might give us a more refined segmentation process.

\begin{figure}[!h]
\begin{adjustwidth}{-2.25in}{0in} 
\includegraphics[width=1.4\textwidth]{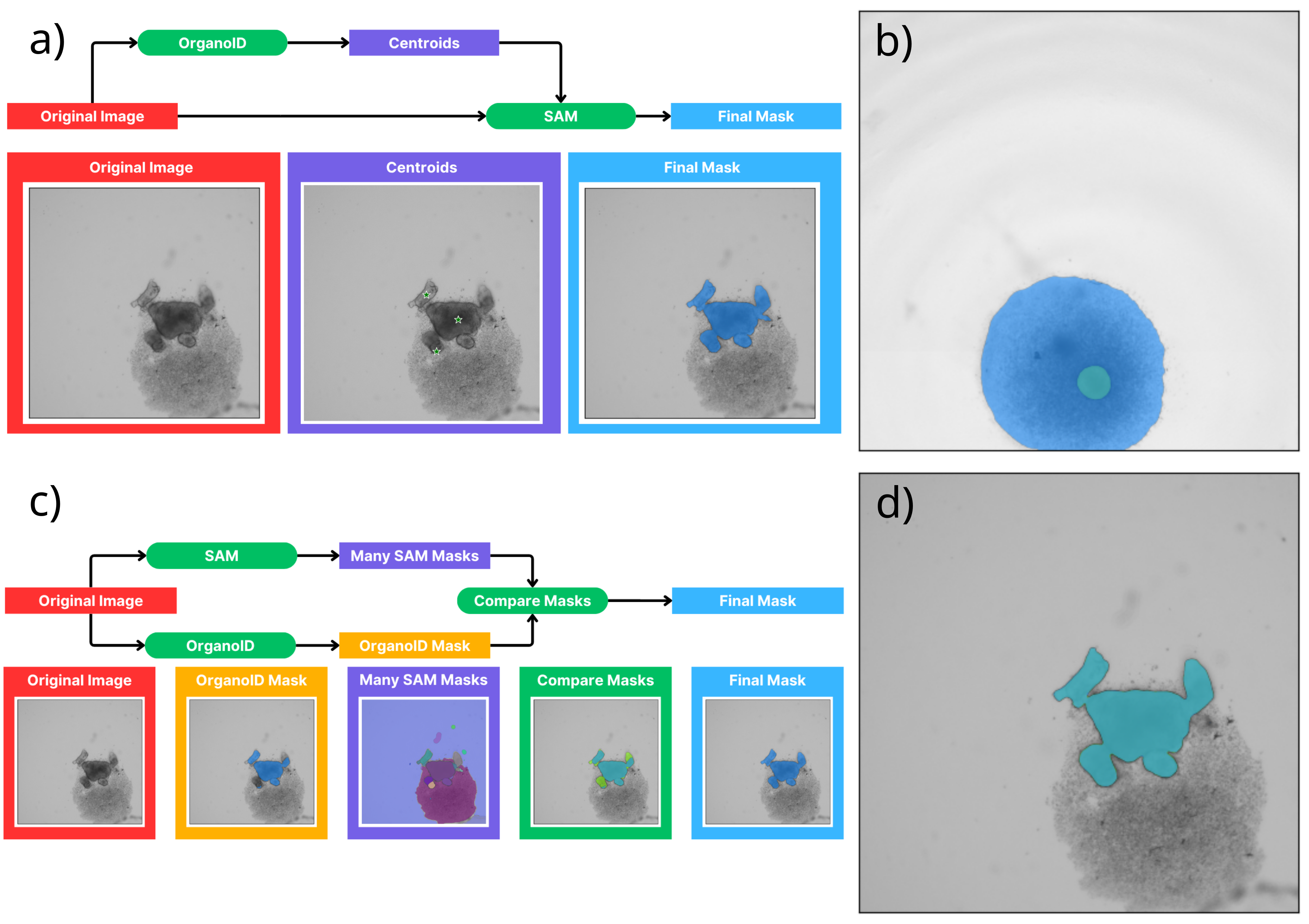}

\caption{{\bf The OrganoID Centroid + SAM and OrganoID + SAM Composite processes} a) Flowchart for the OrganoID Centroid + SAM process. Given the original image, OrganoID produces a mask (which we discard) and outputs several metrics, including the centroid of each distinct organoid in the mask. We then pass the array of centroids, along with the original image, to SAM, which produces a final mask. b) An example mask produced by Trained OrganoID Centroid + SAM from an image in {\bf Image Set C}. The ground truth mask is shown in green, overlaid with the calculated mask in blue. (IOU=0.03) c) Flowchart for the OrganoID + SAM Composite process. The original image is passed both to OrganoID, which produces one rough mask for the organoid, and to SAM, which produces an array of masks representing every distinct object it can identify in the image. We then compare each of these SAM masks to the OrganoID mask, accepting those which match at least part of the OrganoID mask, and rejecting the rest. Assuming the accepted masks to represent portions of the true organoid, we then merge them into a single, final mask. d) An example mask produced by the Trained OrganoID + SAM Composite process from an image in {\bf Image Set B}. The ground truth mask (green) and SAM Composite mask (blue) are overlaid and almost perfectly agree. (IOU=0.99)}
\label{trained}
\end{adjustwidth}
\end{figure}

\begin{figure}[!h]
\begin{adjustwidth}{-2.25in}{0in} 
\centering
\includegraphics[width=.95\textwidth]{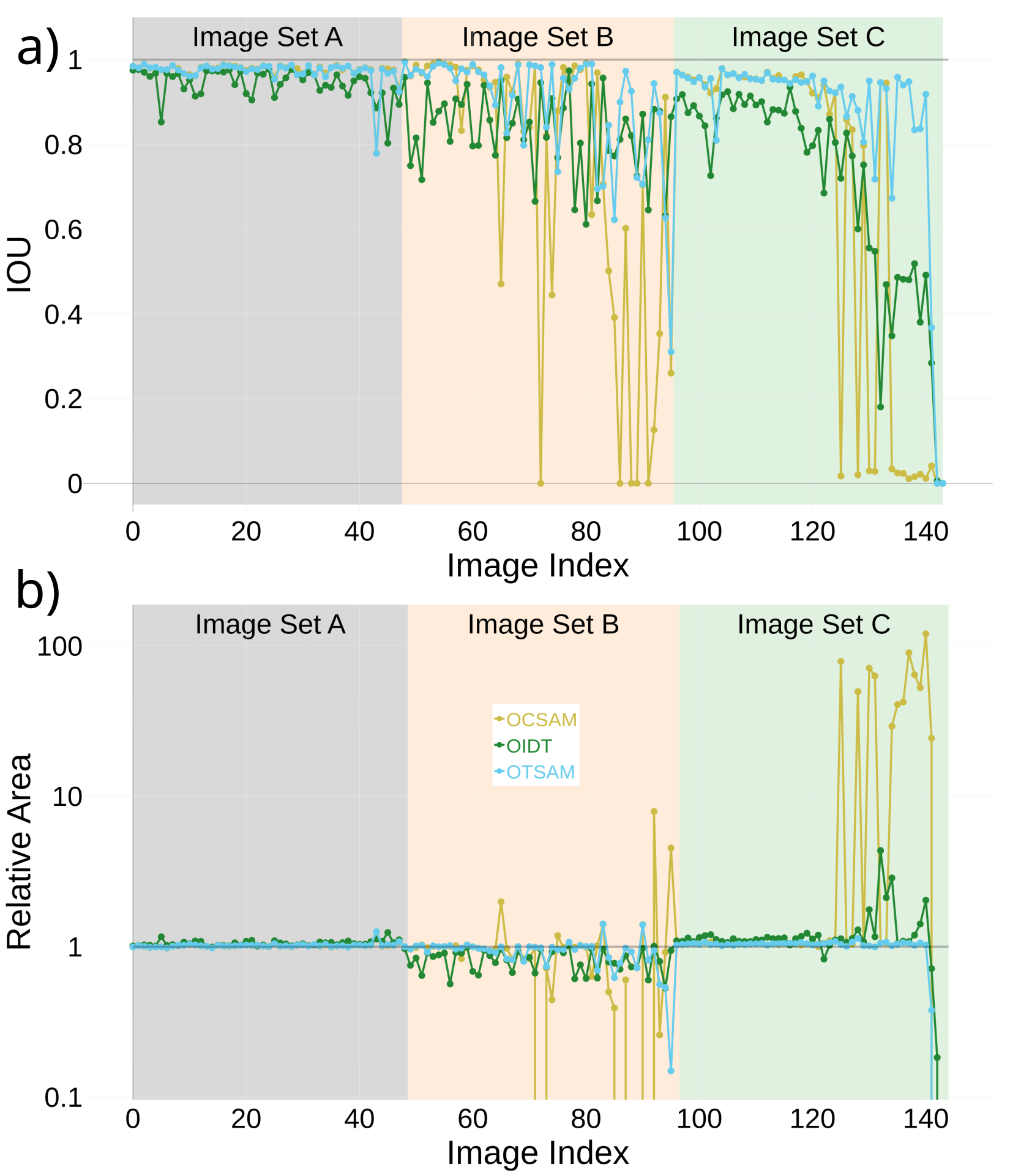}

\caption{{\bf IOUs and normalized areas of the OrganoID Centroid + SAM and OrganoID + SAM Composite processes} From Figs.~\ref{fig:all}a Fig.~\ref{fig:all}b we select the results for the Trained OrganoID (OIDT), Trained OrganoID Centroid + SAM (OCSAM), and Trained OrganoID + SAM Composite (OTSAM) methods for comparison, and show a) the IOU, and b) the relative area. For most individual images, the OIDT Centroid + SAM method does improve the segmentation slightly. But when the masks it generates are worse, they are dramatically worse, so it is worse on average that Trained OrganoID alone, and we can dismiss it as a method. The Trained OrganoID + SAM Composite, however, is better than Trained OrganoID alone on almost every single image, and a dramatic improvement on average.}
\label{traineddata}
\end{adjustwidth}
\end{figure}

We initially developed 2 methods that combine Trained OrganoID and SAM. The first method, which we call the Centroid method (OCSAM), takes the centroid of the Trained OrganoID mask and uses it as the starting point for the SAM segmentation (Fig.~\ref{trained}a). When OrganoID measures the various metrics of the organoids it finds, such as area and eccentricity, it also measures and records the centroid, akin to a "center of mass" of each organoid. We can then take the location of each organoid and feed it into SAM, which will try to segment the object in the image which contains each point, producing a mask of SAM quality that leverages OrganoID's ability to identify organoids. The other method, which we call the SAM Composite method (OTSAM), is more complicated, using SAM to segment every object it can identify in the image (often more than 20), then comparing those masks directly to the mask produced by Trained OrganoID, as shown in Fig.~\ref{trained}c. By combining any masks that agree well with at least a portion of the OrganoID mask, we get a final mask that should match OrganoID overall, and should track details like SAM. Figs.~\ref{traineddata}a, \ref{traineddata}b, and~\ref{trained}b show results for this method. As seen in those figures, the Centroid method often improves the mask, as expected, but the improvement is usually slight, and in a relatively large number of cases, the mask is actually made dramatically worse. In fact, the Centroid method has the same tendency to overshoot the boundaries of the organoid that GD + SAM had. The Composite method, on the other hand, does not have this issue, and turns out to be extremely accurate. It combines the consistency of OrganoID with the precision of SAM, as seen in Fig.~\ref{trained}d, and is by far the most accurate individual segmentation method we found.
	
\begin{figure}[!h]
\begin{adjustwidth}{-2.25in}{0in} 
\includegraphics[width=1.4\textwidth]{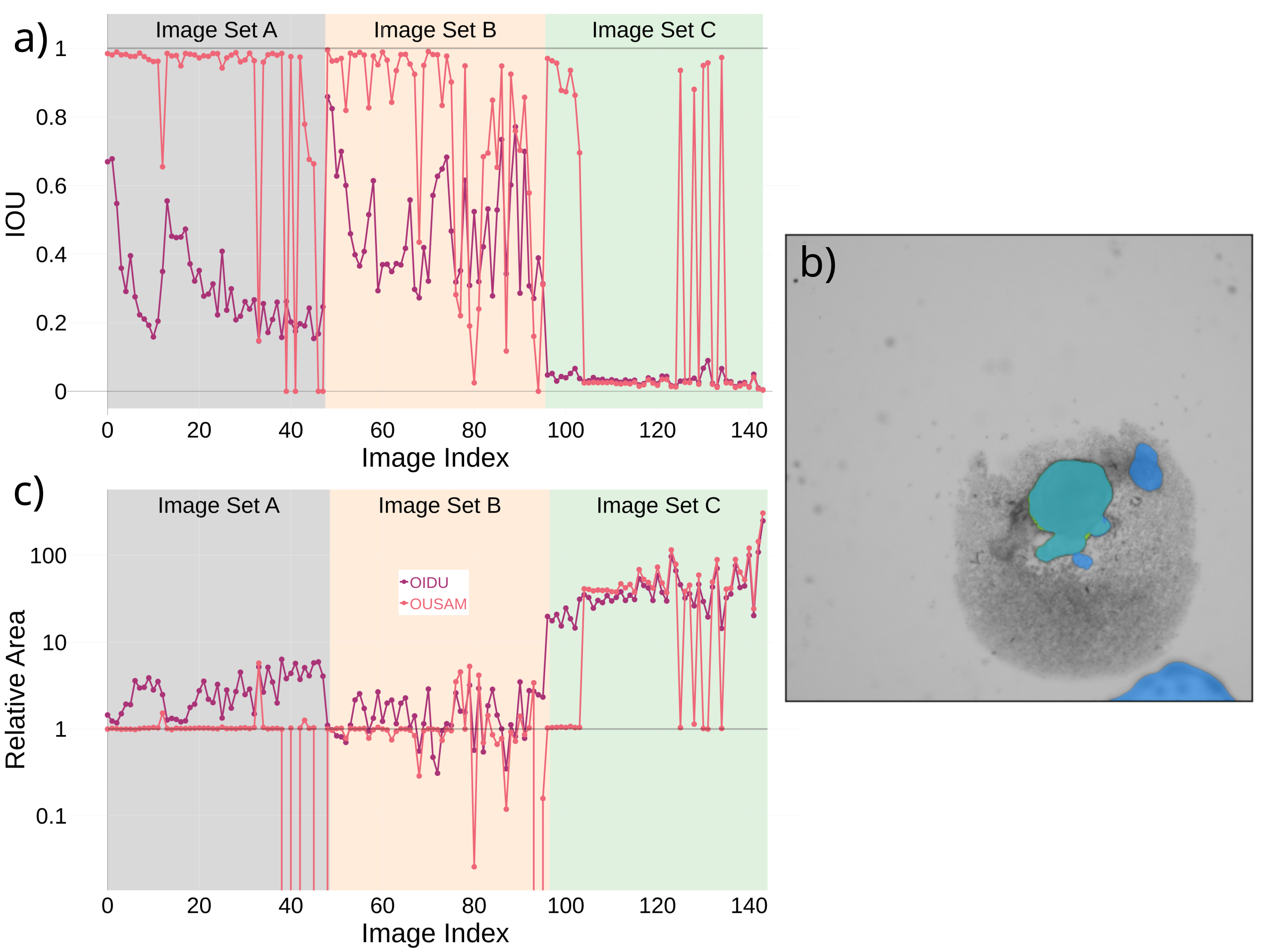}

\caption{{\bf Segmentation methods based on Untrained OrganoID} a) From Fig.~\ref{fig:all}a we select the IOUs from the Untrained OrganoID (OIDU) and Untrained OrganoID + SAM Composite (OUSAM) methods for comparison. Even based on Untrained OrganoID, the Composite method is able to produce some masks that are almost perfect, and does so most of the time in {\bf Image Set A}, less consistently in {\bf Image Set B}, and only a handful of times in {\bf Image Set C}. When it works, it works very well, but it is unreliable when the starting masks are as poor as those produced by Untrained OrganoID. b) An example mask produced by Untrained OrganoID + SAM Composite from an image in {\bf Image Set B} in blue, with the ground truth mask overlaid in green. (IOU=0.58) c) From Fig.~\ref{fig:all}b we select the areas from the Untrained OrganoID and Untrained OrganoID + SAM Composite methods for comparison. In {\bf Image Set A}, Untrained OrganoID has a tendency to overestimate areas. For most images, the Composite method completely corrects this tendency, but in the few cases where it finds no organoid at all, it returns an area of 0, for an overall negative bias. Meanwhile, in {\bf Image Set C}, Untrained OrganoID is consistently and dramatically overestimating the areas of the organoids, and while the Composite method can correct this for about a dozen images, it actually tends to overestimate the area a little bit more than Untrained OrganoID alone on the remaining images.}
\label{untrained}
\end{adjustwidth}
\end{figure}

Having used the Composite method to significantly improve Trained OrganoID, we wondered if we could get a similar degree of accuracy without having to provide manually-generated, experiment-specific training data. In particular, our composite strategy requires an approximate initial segmentation, but it is not a priori known how accurate the approximate mask must be in order for the SAM Composite correction to work. Thus, we combined the same SAM Composite process with the masks generated by Untrained OrganoID to examine its performance with these rougher prototype masks. Fig.~\ref{untrained}a, Fig.~\ref{untrained}b, and Fig.~\ref{untrained}c show that this method does provides significant improvements, but not to the same degree or with the same consistency as retraining did. SAM methods tend to give masks that are either very accurate or completely inaccurate, and the quality of the starting mask determines how often it is correct. Thus, Untrained OrganoID + SAM (OUSAM) often performs comparably to Trained OrganoID + SAM, but is noticeably less consistent, especially on {\bf Image Set C}.

\begin{figure}[!h]
\begin{adjustwidth}{-2.25in}{0in} 
\centering

\includegraphics[width=1.4\textwidth]{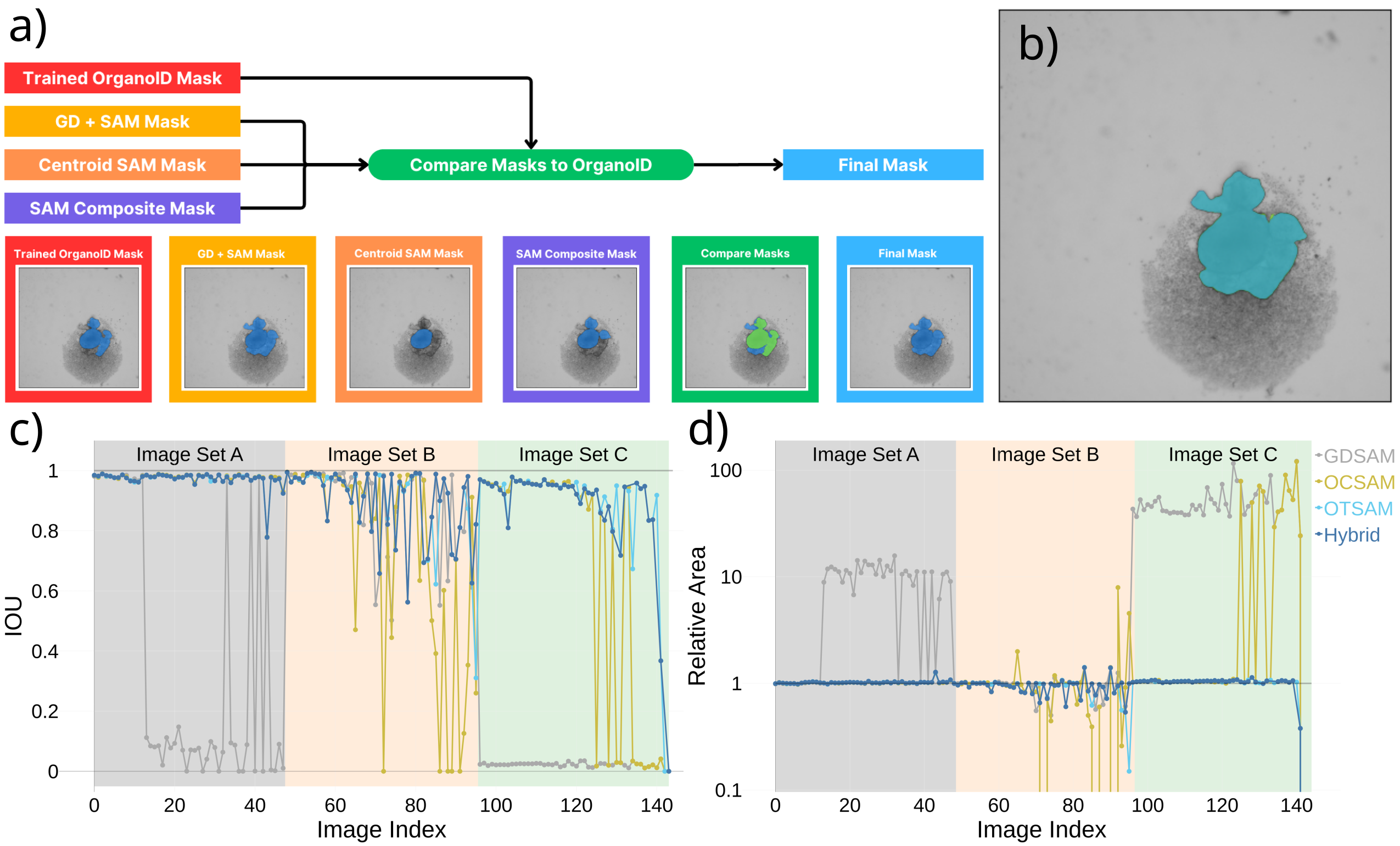}
\caption{{\bf IOUs for the Hybrid method and related methods} a) Flowchart for the Hybrid method. We assume that Trained OrganoID will always be close to the true mask, since it is the least prone to large errors, but another method will usually give a more accurate solution. We compare each of the other masks to the OrganoID mask and select the one that has the greatest overlap with it, and call this our Hybrid mask. b) An example Hybrid mask. Blue represents the Hybrid mask, and green represents the ground truth mask. The two masks are almost identical (IOU=0.99). In this particular example, the Hybrid has correctly identified the GDSAM mask as the best candidate. c) From Fig.~\ref{fig:all}a we select the IOUs from the Hybrid method, as well as Grounding DINO + SAM (GDSAM), Trained OrganoID Centroid + SAM (OCSAM), Trained OrganoID + SAM Composite (OTSAM) methods. The Hybrid method works by taking the masks of the other 3, comparing them to OrganoID, and picking the one that agrees with OrganoID best, so we compare it here with its constituent parts. The Hybrid method does perform better overall than any of its constituent methods, but Trained OrganoID + SAM Composite is by far the best individual method, and the Hybrid method's advantage over it is almost negligible. d) Relative areas for the same methods as in panel c, selected fromFig.~\ref{fig:all}b. Once again, the improvements of the Hybrid method over Trained OrganoID + SAM Composite are marginal.}
\label{hybrid}
\end{adjustwidth}
\end{figure}

Now that we have several segmentation methods that tend to produce different kinds of errors, we can try one more way to improve performance, namely to segment the same image in several ways and compare the results. We have observed that Trained OrganoID almost always gets the general shape and location of the organoid correct, but tends to make small errors around the edges, and methods utilizing SAM can correct this. Thus, we can segment a single image using Trained OrganoID, Trained OrganoID Centroid + SAM, Trained OrganoID + SAM Composite, and Grounding DINO + SAM, and of the 3 masks that use SAM, we select the one that most closely matches Trained OrganoID. Furthermore, if none of the masks match OrganoID well, we do not produce an output. We call this approach the Hybrid method. If we are interested in analyzing individual images, we obviously cannot do this, but in our research we are much more interested in characterizing large sets of images, so excluding a few images from consideration because we think those measurements are unreliable is a perfectly viable strategy. Figs.~\ref{hybrid}c and~\ref{hybrid}d show that the Hybrid method performs almost identically to Trained OrganoID + SAM, except on {\bf Image Set C}, where it is able to exclude a few poorly-segmented images to produce a slightly higher average IOU, which can be seen in Fig.~\ref{meanious}. In practice, this is the Hybrid method's only substantial advantage, namely its ability to skip over images where no two segmentation methods agree, since any mask it could provide for those images would be suspect.
For our purposes, we had already developed all 4 of these protocols, so using something like the Hybrid method only adds compute time, which is worthwhile for even a marginal increase in accuracy. However, for other users, the benefits may often not be sufficient to justify the extra effort in developing and running the required software.

\begin{figure}[!h]
\begin{adjustwidth}{-2.25in}{0in} 
\centering
\includegraphics[width=1.4\textwidth]{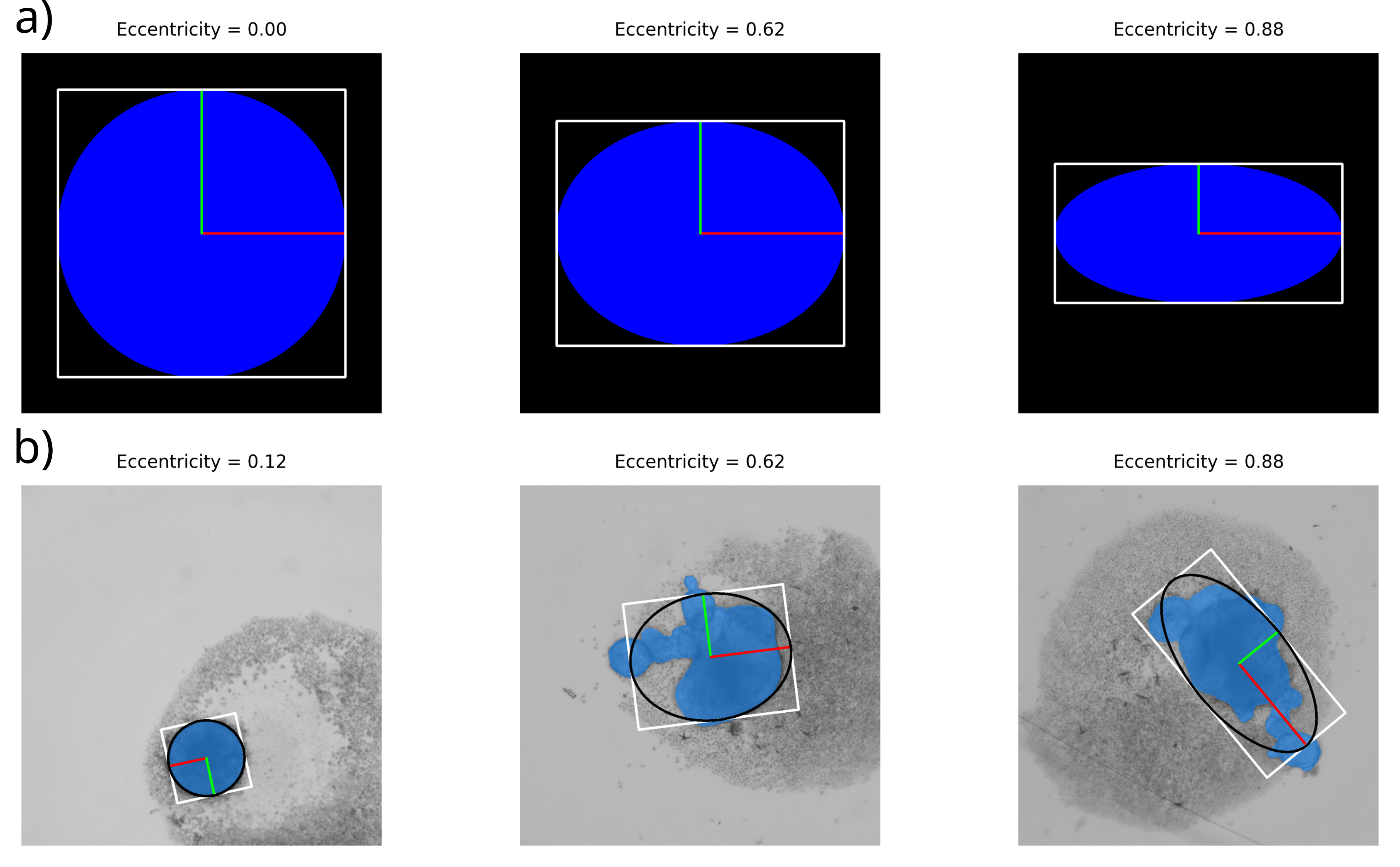}

\caption{{\bf Demonstration of Eccentricity Calculation} a) Eccentricity measures the deviation of an elipse from a circle. An eccentricity of 0 represents a perfect circle, while an eccentricity of 1 would be a perfectly flat shape. Three examples are shown here. b) We define the eccentricity of an arbitrary shape as the eccentricity of an elipse with the same second moments as the given mask. This lets us quantify the eccentricity of irregularly shaped organoids, as shown in three examples here.}
\label{eccentricity_calc}
\end{adjustwidth}
\end{figure}

\begin{figure}[!h]
\begin{adjustwidth}{-2.25in}{0in} 
\centering
\includegraphics[width=1.4\textwidth]{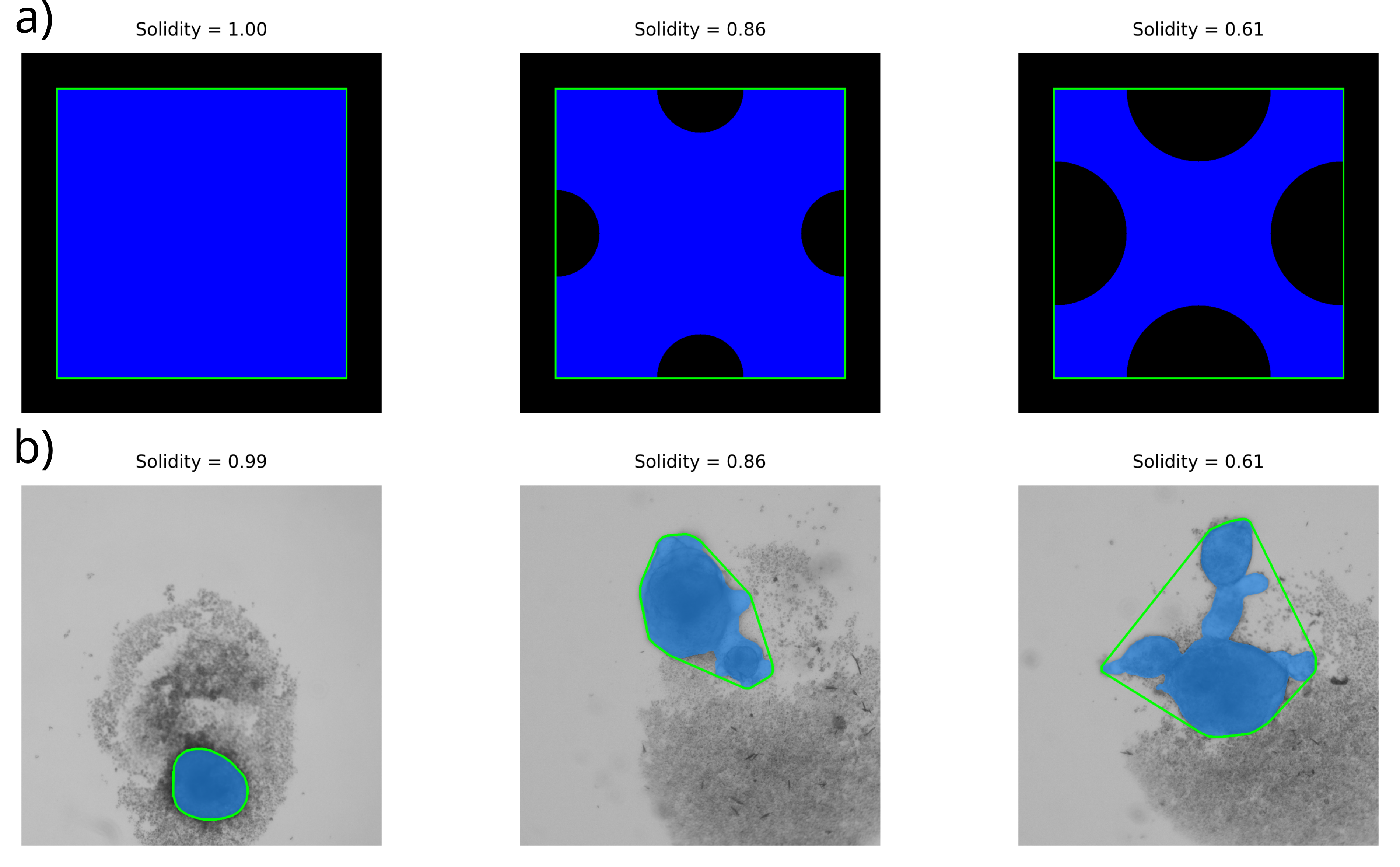}

\caption{{\bf Demonstration of Solidity Calculation} a) The solidity of a shape is the ratio of its area to the area of its convex hull. The convex hull of a shape is the enclosing shape with the smallest possible perimeter, as shown in these 3 examples, which all have the same square-shaped convex hull, outlined by the green perimeter. b) Three example masks with their convex hulls outlined in green and their solidities indicated above each one.}
\label{solidity_calc}
\end{adjustwidth}
\end{figure}

\begin{figure}[!h]
\begin{adjustwidth}{-2.25in}{0in} 
\centering
\includegraphics[width=1.4\textwidth]{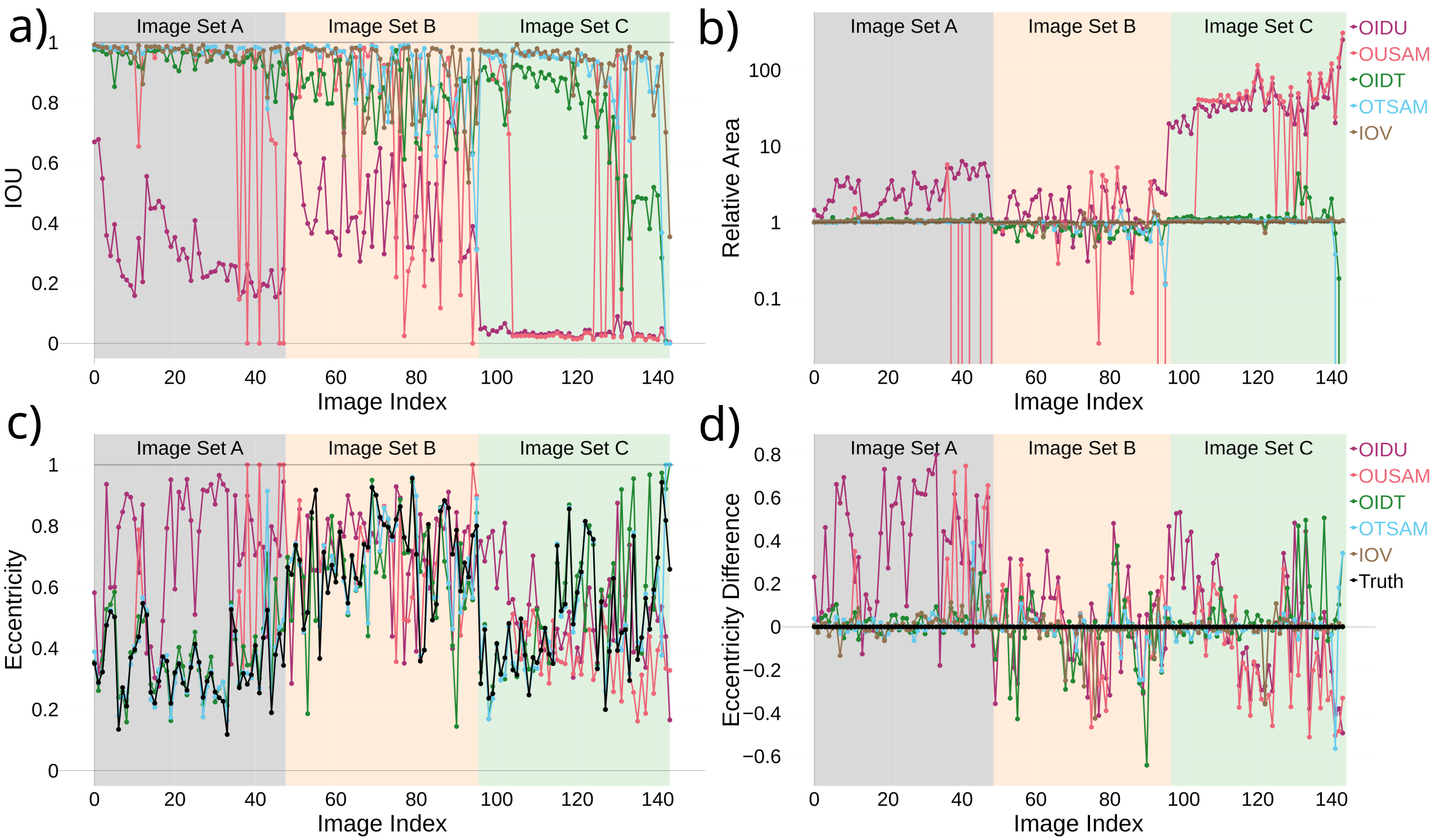}

\caption{{\bf OrganoID and successful enhancements thereof} a) IOU for the Untrained OrganoID (OIDU), Trained OrganoID (OIDT), Untrained OrganoID + SAM Composite (OUSAM), and Trained OrganoID + SAM Composite (OTSAM), selected from Fig.~\ref{fig:all}a. b) Areas for the same methods, selected from Fig.~\ref{fig:all}b. c) Eccentricity calculated by the {\it skimage} Python library, and defined as the eccentricity of an elipse with the same second moments as the organoid mask (see Fig.~\ref{eccentricity_calc}). The horizontal axis and line colors are the same as in all previous plots, except that here we plot the ground truth data in black. In previous plots, we could omit this, since the IOU and relative area of a ground truth mask are both 1.0 by definition. Note that when a segmentation method returns an empty mask, we plot that as an eccentricity of 1, which is impossible otherwise. The true eccentricity is low in {\bf Image Sets A} and {\bf C}, since the organoids are quite round, and Untrained OrganoID is particularly likely to significantly overestimate the eccentricity in {\bf Image Set A}, producing an erroneous mask with a very irregular shape. (Inter-observer variation omitted from this plot to avoid eccessive visual clutter.) d) Difference between the measured eccentricity shown in panel c and the true ground truth eccentricity for the same image. 
}
\label{big4}
\end{adjustwidth}
\end{figure}

\begin{figure}[!h]
\begin{adjustwidth}{-2.25in}{0in} 
\centering
\includegraphics[width=1.4\textwidth]{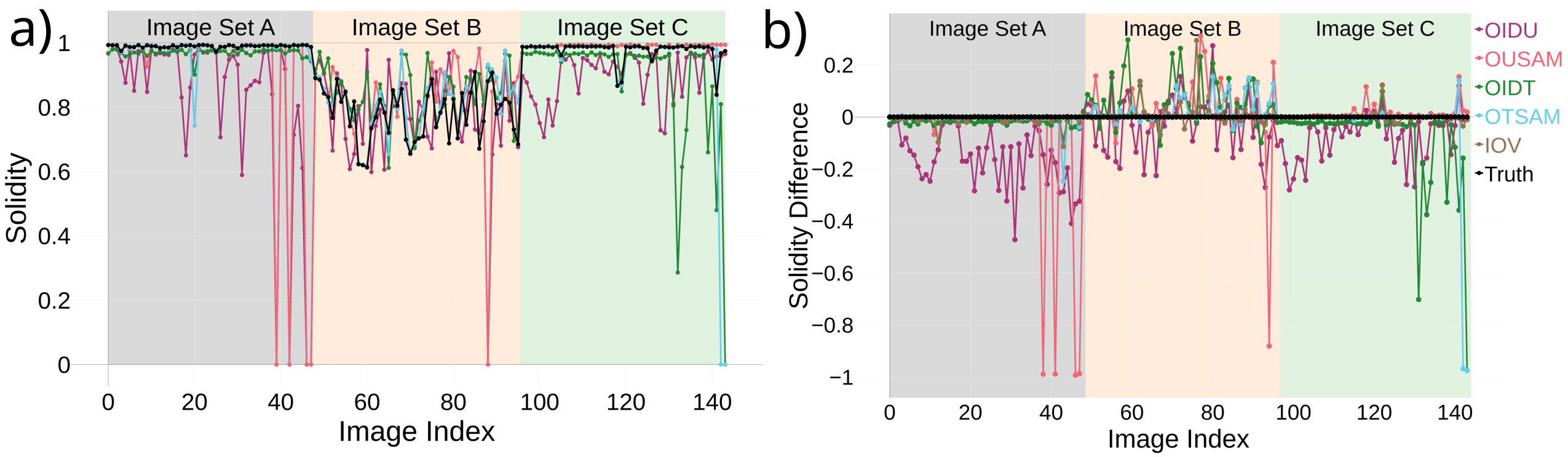}
\caption{{\bf OrganoID and successful enhancements thereof (continued)} a) Solidity calculated by the {\it skimage} Python library, and defined as the ratio of a mask's area to the area of its convex hull (see Fig.~\ref{solidity_calc}). The horizontal axis and line colors are the same as in Fig.~\ref{big4}c. In this case, an empty mask is considered to have solidity 0, which is once again impossible otherwise. b) Difference between the measured solidity shown in panel a and the true ground truth solidity for the same image. (Inter-observer variation omitted from this plot to avoid eccessive visual clutter.) Since solidity ranges from 0 to 1, and most ground truth solidities outside of {\bf Image Set B} are almost exactly 1, almost all errors are negative in the other two image sets. On {\bf Image Set B}, however, the errors tend to be, if anything, more often positive, meaning that the segmentation errors consist of cutting off protrusions of the organoid, making a mask more rounded and regular than the true shape. This is consistent with Fig.~\ref{big4}b, where there was a slight negative bias in the areas on {\bf Image Set B}.}
\label{big4cont}
\end{adjustwidth}
\end{figure}

The SAM Composite method has provided by far the best results, to this point, and we have applied it using two initial segmentation strategies, Trained and Untrained OrganoID. We further examine the effectiveness of the SAM Composite process by comparing these 4 segmentation methods (OIDU, OIDT, OUSAM, and OTSAM) using IOU (Fig.~\ref{big4}a), relative area (Fig.~\ref{big4}b), eccentricity (Figs.~\ref{big4}c and~\ref{big4}d), and solidity (Figs.~\ref{big4cont}a and~\ref{big4cont}b) as our metrics. Eccentricity and solidity both measure the deviation of organoids from a perfectly round shape. This often represents an aberrant morphology, corresponding to diseased or otherwise unhealthy cells. For instance, in our experiments, we found that high levels of CEPT promoted the formation of large, extremely round spheroids, while no CEPT led to no spheroid formation at all. Meanwhile, marginal doses of CEPT, just high enough to form spheroids, led to the highest eccentricities and lowest solidities, whereas a perfectly healthy, circular spheroid would have a solidity near 1 and eccentricity close to 0. This was especially the case in cell line 2, as seen in {\bf Image Set B} in Figs.~\ref{big4} and \ref{big4cont}. The exact methods of calculating eccentricity and solidity are shown with examples in Figs.~\ref{eccentricity_calc} and \ref{solidity_calc} respectively. Note that for eccentricity and solidity, we include the ground truth values as a 5th segmentation method for comparison, since the ground truth values actually vary, unlike IOU and relative area, which are 1 by definition. For eccentricity, note that the ground truth eccentricity is significantly higher for {\bf Image Set B} than for {\bf A} or {\bf C}, which is why it is actually more difficult to overestimate it within the finite range of 0 to 1. However, since the ground truth varies so much, it is difficult to actually compare the accuracy of various methods, so in Fig.~\ref{big4}d, we subtract the ground truth eccentricity from the eccentricity measured by each method and plot the difference. Fig.~\ref{big4cont}a is clearer in {\bf Image Sets A} and {\bf C}, since the true solidity is so close to 1, but in {\bf Image Set B}, the true solidity varies enough to make it similarly difficult to read, and so we plot the difference in Fig.~\ref{big4cont}b the same way we did in Fig.~\ref{big4}d. Finally, in Fig.~\ref{big4}a, b, and d and Fig.~\ref{big4cont}b, we added an independent human segmentation for comparison. These additional segmentations were performed in the same manner as the ground truth masks, except generated independently by HC, rather than GG and CC.

As noted before, training dramatically improves the accuracy of the segmentation, measured both by IOU (Fig.~\ref{big4}a) and relative area (Fig.~\ref{big4}b), and the improvement is very consistent. In only 3 images out of 148 does the untrained version of OrganoID have a better IOU than the trained one, and only by a small margin in those cases. From Figs.~\ref{big4}c and~\ref{big4cont}a we can see that training similarly improves eccentricity and solidity measurements, though the improvement is most clear on {\bf Image Sets A} and {\bf C}. Note that any round mask will have very low eccentricity and high solidity, so even in cases where we can see from Figs.~\ref{big4}a and~\ref{big4}b that the masks are far from correct, for instance with OIDU + SAM on {\bf Image Set C}, the eccentricities and solidities in Figs.~\ref{big4} and~\ref{big4cont} would give no such indication. Solidity, in particular, presents a difficulty for {\bf Image Sets A} and {\bf C}. The true solidity for these image sets is seldom less than 0.98, but even trained OrganoID rarely reports a value greater than .97. Because the variance between images is so little, the precision necessary to distinguish between them is beyond OrganoID alone. However, the solidity measurements of Untrained OrganoID are too inaccurate to distinguish even the difference between the irregular organoids of {\bf Image Set B} and the very regular, round ones of {\bf A} and {\bf C}, whereas Trained OrganoID shows this difference clearly, even if it tells us little about the differences within an image set.\\
The SAM Composite method, meanwhile, improves the accuracy of both IOU and area (Figs.~\ref{big4}a and~\ref{big4}b), but not with the consistency of retraining. If applied to untrained OrganoID, it tends to make individual image segmentations either dramatically better (most images in {\bf Set A}) or somewhat worse (most images in {\bf Set C}). It also tends to dramatically overestimate areas in {\bf Image Set C}, as noted previously, which retrained OrganoID does not, with or without SAM. Note that sometimes the Untrained OrganoID + SAM Composite process returns no mask at all, which results in an IOU of 0, area 0, eccentricity 1, and solidity 0. Because of the smooth edges of SAM Composite masks, we can see in Fig.~\ref{big4cont}a that both Untrained and Trained OrganoID + SAM Composite show the same uniformly high solidity as ground truth in {\bf Image Sets A} and {\bf C}, even if we can see from Figs.~\ref{big4}a and~\ref{big4}b that the masks themselves are not at all accurate for those images. The most notable mistakes are when Untrained OrganoID + SAM Composite returns no mask at all for a solidity of 0. In {\bf Image Set B}, where the solidity is nontrivial, the SAM Composite method and retraining both improve accuracy, and the two combined are the most accurate. Finally, Fig.~\ref{big4}d shows that retraining is the most important improvement, and that, once again, both combined give the greatest accuracy.

\begin{figure}[!h]
\begin{adjustwidth}{-2.25in}{0in} 
\centering
\includegraphics[width=1.4\textwidth]{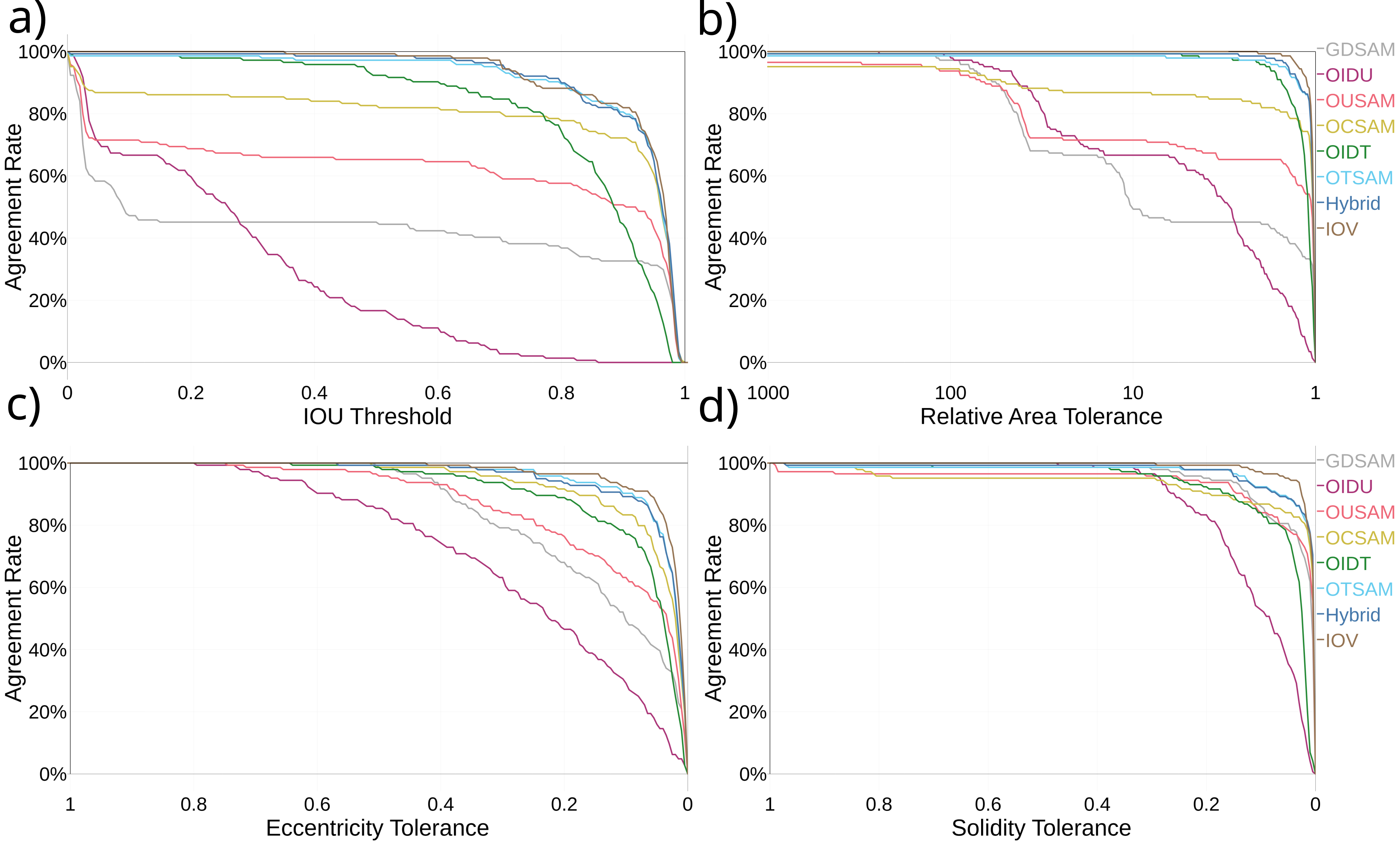}

\caption{{\bf Agreement of segmentation methods with ground truth segmentation} Here we show the rate at which each segmentation method agrees with manual segmentation, given a certain threshold or tolerace for correctness. a) The x-axis gives a criterion in the form of a minimum IOU, and the y-axis shows the percentage of masks produced by a certain method which have at least as great an IOU with the ground truth mask. By this metric, discrepancy between the OTSAM or Hybrid method and the ground truth is at the level of inter-observer variability (IOV). b) The x-axis gives a tolerace in the relative area, and the y-axis shows the percentage of masks with a relative area $R$ such that $\frac{1}{x} \leq R \leq x$. (Note that, in this scheme, a tolerace of 1 represents perfect agreement.) c) The x-axis represents a tolerance in eccentricity and the y-axis gives the percentage of masks produced by each method such that the absolute value of the eccentricity difference is less than the given tolerace. d) The x-axis represents a tolerance in solidity and the y-axis gives the percentage of masks produced by each method such that the absolute value of the solidity difference is less than the given tolerace.
}
\label{agreement}
\end{adjustwidth}
\end{figure}

Finally, we here present plots to rank segmentation methods by error rate, given some acceptable threshold for errors. In each subplot of Fig~\ref{agreement}, we select one metric of interest and show what percentage of masks generated by each segmentation method give the "correct" value for that metric for any given threshold of correctness. The most striking result is in Fig.~\ref{agreement}a, which shows that, for any given IOU with the ground truth, the most direct measurement of mask correctness, our best automated segmentation methods perform at the same level as inter-observer variability (IOV), i.e., the discrepancy between a human and the OTSAM or Hybrid method is equivalent to the discrepancy between two human segmenters. This indicates that OTSAM and the Hybrid method are performing as well as any automated method can, since an automated method that outperforms IOV would be overfitting, capturing the biases and subjective judgements of the single individual who segmented the training data. In the other metrics, particularly solidity, OTSAM and Hybrid do still slightly underperform IOV at the smallest tollerances. As for the other metrics, Fig.~\ref{agreement}a and b show certain methods (GDSAM, OCSAM, and OUSAM in particular) plateauing at intermediate error tolerances and thresholds. These are the weaker SAM-based methods, and this recapitulates the bimodal behavior observed earlier, where SAM can produce very high-quality segmentations, but paired with an unreliable identification tool, it does this very inconsistently.

\section*{Conclusion}
We have examined several possible methods for automated segmentation of organoid microscopy images. We have shown that OrganoID in its original form does not perform optimally with organoid image data showing misshapen organoids, presence of dead cells and/or debris, unfavorable illumination conditions, or other confounding factors. We found that its accuracy can be improved by retraining its model weights using more specific training data, but the results are still not accurate enough for reliable automated analysis. We also examined the open-source segmentation tool SAM, which, combined with GroundingDINO, can also segment organoid images, but performs poorly, with a similar accuracy to untrained OrganoID, albeit with dissimilar errors. 

To produce more accurate segmentations, we developed a composite method that uses SAM to form a composite mask, based on a preliminary segmentation by OrganoID. This enhancement provides a dramatic increase in accuracy, finally producing segmentations that can be relied on for automated analysis even when a non-negligible percentage of the images are less than optimal. 

We also tried a similar composite method combining SAM with the untrained version of OrganoID, but the results were inconsistent, showing that there is a threshold of accuracy that the preliminary segmentation method in the combination must reach in order for the compositing to be able to significantly improve it. We then showed that combining several segmentation methods may result in a slightly more consistent segmentation protocol, as seen with the Hybrid method. However, the improvements we observed from the Hybrid method were minor enough in comparison to the extra effort required to make the method not worthwhile in most cases.

Finally, we compared the accuracy of all of our methods to the inter-observer variability (IOV) between independent human annotators. We found that the accuracy of our retrained OrganoID + SAM Composite method performed very near the level of IOV. In particular, when measured using intersection-over-union (IOU), the most comprehensive metric, the agreement with ground truth was at the level of inter-observer variability. 

\section*{Supporting information}

\paragraph*{S1 Images}
\label{S1_Images}
The microscopy images used for this work are available at https://doi.org/10.5281/zenodo.19961879.

\paragraph*{S2 Code}
\label{S2_Code}
The GitHub repository containing the code we developed can be found at https://doi.org/10.5281/zenodo.20027217.

\section*{Acknowledgments}
We gratefully acknowledge the donation by Tapawize.ai of a Lambda Vector computer used for this research. CC acknowledges partial support from the same company.

\bibliography{Image-segmentation-organoids_Cartwright-etal.bib}

\end{document}